**DODA: A Database of Datasets for Aesthetics Research**

Lisa Koßmann*[1, 2], Ralf Bartho*[3], Christoph Redies[3] and Johan Wagemans[1,2]

[1] Laboratory of Experimental Psychology, Department of Brain and Cognition, University of Leuven (KU Leuven), Leuven, Belgium

[2] Leuven AI Institute, University of Leuven (KU Leuven), Leuven, Belgium

[3] Experimental Aesthetics Group, Institute for Anatomy I, University Hospital Jena, Germany

*Lisa Koßmann and Ralf Bartho are shared first authors.

**Author Note**

Lisa Koßmann https://orcid.org/0000-0003-3246-7214
Ralf Bartho
Christoph Redies https://orcid.org/0000-0002-5220-8319
Johan Wagemans https://orcid.org/0000-0002-7970-1541

This work is funded by an ERC Advanced Grant (No. 101053925) awarded to JW.
We have no conflicts of interest to disclose.
Correspondence concerning this article should be addressed to Lisa Koßmann, KU Leuven, Tiensestraat 102 - bus 3711, 3000 Leuven, Belgium
Email: lisa.kossmann@kuleuven.be

**Abstract**

With rapid growth in the fields of empirical and computational aesthetics we have seen a vast increase in large image datasets annotated for aesthetics. As the image databases differ widely in many respects (e.g., different standards for annotation), it can be tedious to find the dataset that fits one's research needs best. The absence of a centralized open-science search system causes additional problems. Currently, researchers typically share dataset links in papers or on diverse platforms like OSF, GitHub or Dropbox. Manually searching for details like image quality and content often requires downloading all datasets. Therefore, we present the Database Of Datasets for Aesthetics (DODA), an intuitive Web application in which researchers can browse all important datasets for aesthetics research. DODA provides general information about these datasets (size, resolution, type of annotation, number of annotators, etc.) and for many of them also precomputed quantitative image properties. We discuss relevant criteria for selecting a suitable dataset with DODA and illustrate the benefits of reusing datasets. Our approach facilitates collaboration across the fields of empirical and computational aesthetics.

**DODA: A Database of Datasets for Aesthetics**

Aesthetics is a multifaceted field of study. Broadly speaking it encompasses the study of (initially non-aesthetic) determinants, which can influence valorization, as well as aesthetic outcomes, which we can use to “measure” this valorization. We can further divide the so-called aesthetic determinants into objective and subjective ones (Brielmann & Pelli, 2018; Chamberlain, 2022). Objective determinants are stimulus-based, like quantitative image properties (Bartho et al., 2023; Geller et al., 2022; Lyssenko et al., 2016; Redies et al., 2025), including color (Nascimento et al., 2021; Palmer et al., 2013; Schloss & Palmer, 2011; M. Sun & Ying, 2023), symmetry, and balance (Makin et al., 2016, 2018; Hübner & Fillinger, 2016) (Chamberlain, 2022). However, these objective determinants are not always successful in predicting how stimuli are perceived. Subjective determinants take the observer into account, as is the case for subjective order and subjective complexity (Koßmann, Hellemans, et al., 2026; Van Geert et al., 2022, 2025; Van Geert & Wagemans, 2020). Context is also important here, meaning the context of the stimulus (e.g., an image displayed in a museum versus laboratory) and the context of a person (e.g., experience, personality, other demographics) (Chamberlain, 2022). These subjective determinants are often assessed through participant ratings.

There are some determinants, which would ideally be objective, but are hard to measure and operationalize, for example, fluency (Lin et al., 2025; Reber et al., 2004) and composition (Arnheim, 1954; Hook & Glaveanu, 2013; Koßmann, Meulemans, et al., 2026; McManus et al., 2011). Perceptual fluency is often measured by complexity. By contrast, the concept of conceptual fluency is much more subjective because it depends on the viewer’s conceptual framework. Similarly, composition includes the notions of rule-of-thirds for photography and balance, which can be measured objectively by computing the distribution of luminance weight, (and other variables like contrast, color, size, etc.), as well as also more subtle aspects as discussed by Arnheim (Arnheim, 1954; McManus et al., 2011; Hübner & Fillinger, 2016).

Aesthetic outcomes can be split into overall outcomes and specific outcomes. Overall aesthetic outcomes are, for example, appreciation, preference, pleasure, beauty, liking and interest. Multiple models have been proposed to explain the emergence and interplay of these outcomes (see Chatterjee & Vartanian, 2016; Graf & Landwehr, 2015, 2017; Jacobsen, 2006; Leder et al., 2004; Leder & Nadal, 2014; Pelowski et al., 2017; Redies, 2015). As these outcomes are often also assessed through ratings, it is important to note that the way in which participants engage with terminology differs between people and scales. For example, laypeople tend to rely more on affective responses (like Pleasure) than cognitive aspects (like Interest) when making comparative judgements across paintings (Cupchik & Gebotys, 1990). Augustin et al. (2012) found an interplay between generality and specificity in aesthetic word usage in an annotation task for different visual aesthetic object classes.

Specific aesthetic outcomes would be aesthetic emotions, for example, the awe or soothingness that a stimulus elicits. Menninghaus et al. (2019) base their criteria for aesthetic emotions on the work by Kant (1790) and, among other criteria, recognize them as distinct from other types of emotions and as defined either by adding aesthetic meaning to emotion words or by adding emotional meaning to aesthetic terms. The existence of aesthetic emotions is subject of debate; some authors reject this notion on grounds of lacking neural and psychological evidence for emotional responses that are uniquely aesthetic (Kenett et al., 2023; Skov & Nadal, 2020).

The aforementioned distinctions are not trivial, as they approach aesthetics from different angles, each with their own measurements and theoretical framework. In this vast landscape, it can be difficult to find stimulus sets which not only match the types of stimuli one wishes to employ but also match the desired aesthetic outcomes or determinants. Furthermore, the method of measurement might differ between studies. While subjective determinants and overall outcomes are often measured using self-report scales (e.g., Likert scales or visual analogue scales), preference tasks or ranking tasks may also be used. Aesthetic emotions can also be measured indirectly by physiological markers, such as skin conductance, goosebumps, shivers, heart rate, and blood pressure (Kenett et al., 2023).

In recent years, two closely related disciplines have shaped our understanding of aesthetic appreciation: empirical and computational aesthetics (Brachmann & Redies, 2017). Although there is overlap between the two fields (and other terms have been used as well), here we distinguish them as follows. On the one hand, empirical aesthetics aims to understand peoples' aesthetic preferences by conducting empirical research, usually presenting small sets of images to large samples of participants. The approach is often experimental, comparing different conditions (e.g., different types of images) that are assigned (quasi-)randomly to different participants. Nadal and Vartanian (2022) provide an extensive review of studies belonging to empirical aesthetics. Computational aesthetics, on the other hand, is focused on a computational analysis of specific image properties, usually but not necessarily, in relation to the obtained aesthetic preferences or ratings (e.g., with correlations between the different measures or by comparing the image statistics between different conditions). This tradition typically uses very large sets of images (which are needed for machine learning) but very small samples of participants (e.g., a handful). This is probably one of the main reasons why results from models trained on different datasets hardly generalize (Bartho et al., 2023). For a review of studies belonging to computational aesthetics, see Zhang et al. (2021). From this characterization, it is clear that both fields employ datasets of images, annotated with aesthetic ratings or other measures of aesthetic appreciation. In their quest to study isolated determinants of aesthetics, studies in empirical aesthetics have also employed simplified stimuli (e.g., stimuli reduced in complexity and variability) under well-controlled experimental conditions. In contrast, computational aesthetics has mostly used rich, multidimensional stimuli that exhibit too much variability to effectively determine factors relevant for aesthetic appreciation.

Currently across both fields, almost every study is conducted on a novel dataset, with different stimuli and measures, and the full potential to collaborate and generalize findings is not realized. In an attempt to bridge the gap between the two fields, as well as to provide researchers with a tool to easily find and filter relevant datasets for specific study purposes, we are introducing DODA, the Database Of Datasets for Aesthetics. DODA is available online as part of the Aesthetics Toolbox[1] (Redies et al., 2025).

Reusing a dataset is not only efficient, but it also allows for the comparison of different variables across the same stimulus set, enabling a broader and deeper understanding of the factors contributing to aesthetic appreciation. Naturally, it is not always possible to answer one's research question with a pre-existing dataset. In no way do we mean to discourage scientific originality. We simply want to motivate researchers to reflect on whether a new dataset is necessary. Moreover, we want to encourage researchers to share their datasets, whenever possible. A notable example of this practice is the OASIS-Beauty dataset by Brielmann and Pelli (2019) who, in a large data collection effort, added beauty ratings to the open-access Open Affective Standardized Image Set (OASIS) by Kurdi and colleagues (2017). In doing so, they were able to investigate the relation between beauty, valence, and arousal (Brielmann & Pelli, 2019).

[1] https://aesthetics-toolbox.streamlit.app/DODA

**Figure 1**

*DODA Logo*

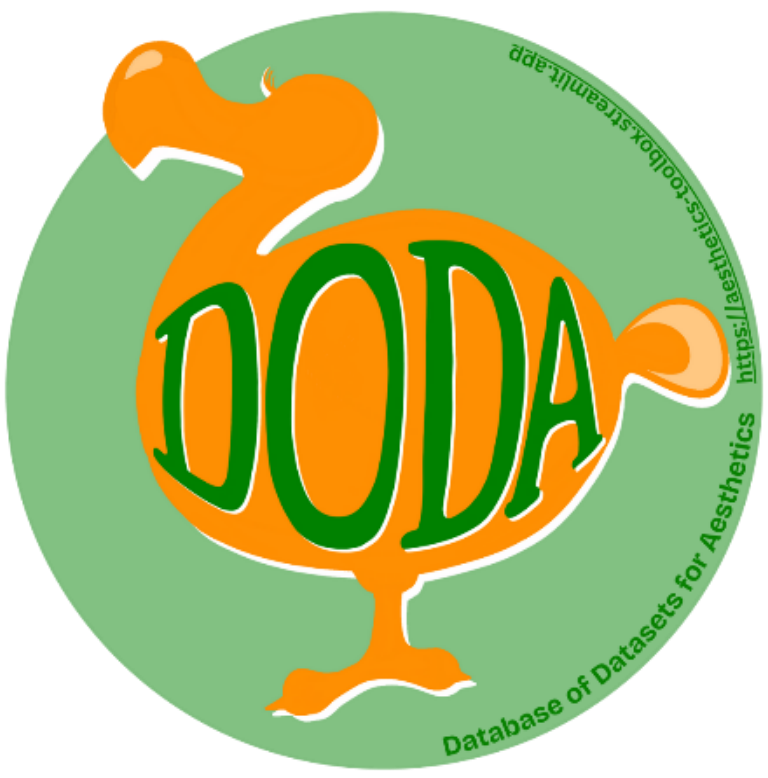

Note: DODA is short for Database of Datasets for Aesthetics. Logo Design Credit: Dr. Nicole Ruta

As all available image databases vary on several dimensions (such as type of images, size of the set, size of the sample, type of annotations, type of measures), it can be quite difficult to find the dataset which best fits one's research needs. Another challenge in finding images is the absence of a centralized Open-Science search system. Currently, researchers typically share dataset links in papers or on platforms like GitHub or Dropbox, leading to manual searches for details like image quality and content, often requiring a download of the entire dataset. On GitHub, some users have collected overviews of image sets (Huckle, 2020a; Liu et al., 2023) and seminal papers in the field of computational aesthetics, providing a great first point of entry, but these do not include smaller datasets from the empirical aesthetics tradition, they have no filter functions and they have less extensive dataset descriptions. Therefore, we created the online application DODA, in which we collect important image datasets for aesthetics and their metadata. Notably, we include image sets of all kinds, paintings as well as photographs and artificially created stimuli, from all ranges of aesthetic quality and all ranges of cultural origin, if they feature an aesthetic annotation. We also include image sets of paintings and photographs of artistic nature that are annotated for a subset or combination of artist, style or genre, as we believe these form a suitable basis for additional annotations, and these are often used in computational aesthetics.

DODA currently encompasses over 120 datasets, published until December 2025 (see Table A in Supplementary Materials) and evaluated on multiple criteria. For each dataset (as applicable) we report: date of publication, number of images, total number of raters, number of raters per image, number of stimulus classes/semantic categories, style of the images, overall content, available annotations, scales of scores, image sources, resolution, field of most common use, task, a link to retrieve the dataset, and the APA style citation of the publication featuring the dataset first. The information is displayed in a table allowing the user to easily skim through and select the criteria most important for their study. DODA is part of the Aesthetics Toolbox, which means that in addition to reviewing available image sets using DODA, users can also easily resize images, or compute quantitative image properties, making the Aesthetics Toolbox an attractive first point of contact for research on digital images.

## Introducing the Database

DODA is a continuously growing effort to facilitate exchange and collaboration among researchers of aesthetics as well as an introduction into the research field. Through conversations with other researchers and our own experience we have identified common checkmarks for assessing whether an image set could be useful for one's research question. Based on a thorough literature review we compiled a list of over 90 datasets and created a short profile for each of them, in the form of a DODA table entry. Below, we will introduce the information available within DODA, as well as the reasoning behind the choice of the most relevant aspects, sometimes highlighted with exemplary entries.

### Year, Author and Citation

For each DODA entry, we provide basic information on its provenance. The reports on year and author refer to the publication, in which the dataset was introduced first. Please note that this entry neither lists the year in which the images were taken or collected nor the artist, photographer or programmer of the images. This point should be observed when updates to datasets are published later. Our collection nicely illustrates the recent growth in the publication of datasets in the field, featuring 14 datasets from 2022 alone.

**Figure 2**

*Publication Count of Image Sets*

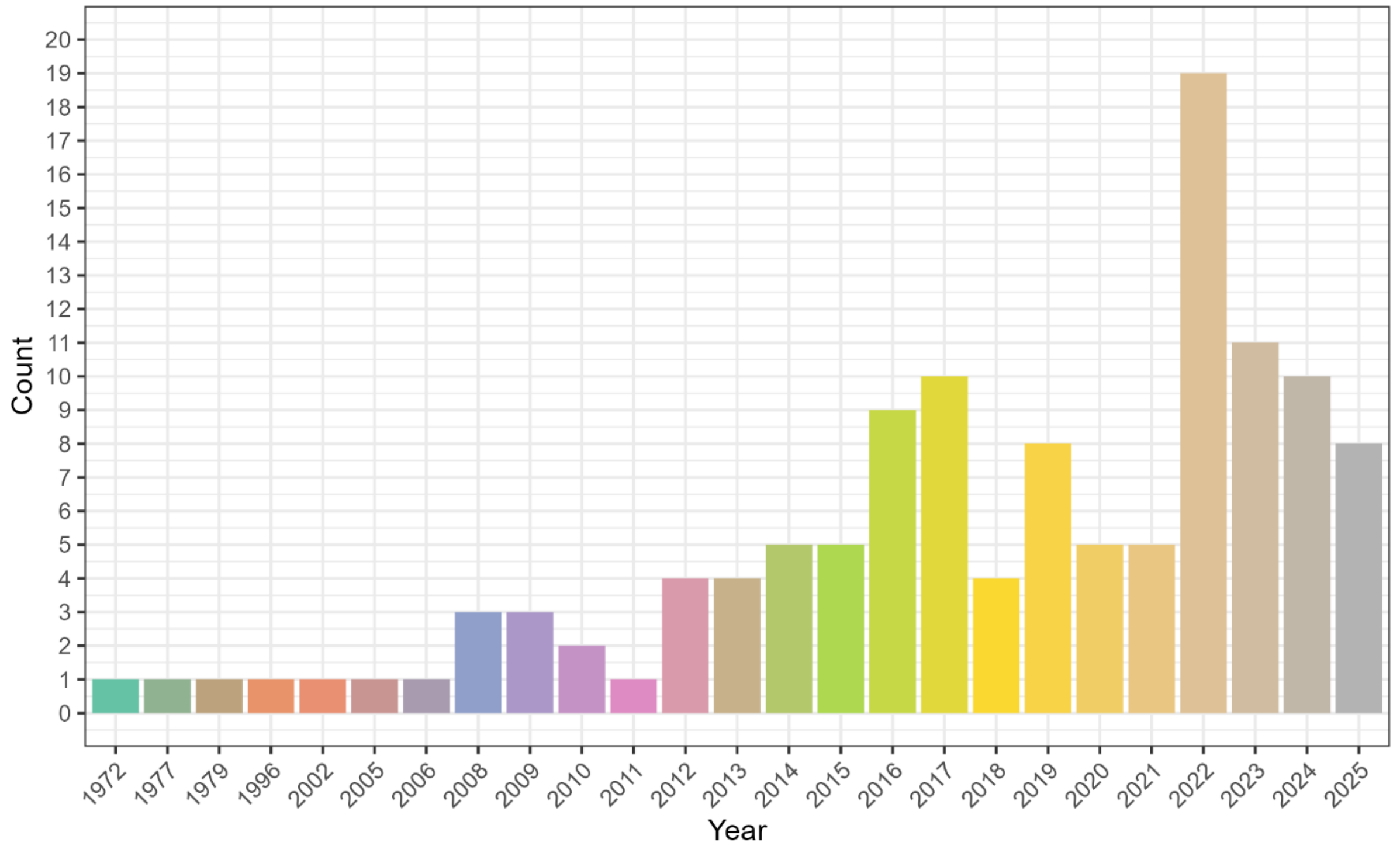

*Note*. Publication Count of Image Sets with Aesthetic Annotations in DODA, plotted with year of publication on the x-axis and count on the y-axis.

**Number of Images**

The number of images in a set is possibly the most important information found in DODA. Users can see at a glance whether a given dataset is large enough for their purposes, or large enough to create subsamples from the dataset. Our largest featured dataset, LAION-Aesthetics (LAION Project, 2022) contains 120,000,000 images. Our smallest dataset, Eisenman & Grove Polygon Preference and Creativity Study (Eisenman & Grove, 1972), contains 12 images. The two exemplary datasets nicely illustrate the potential differences in dataset size between the fields of empirical aesthetics (few images) and computational aesthetics (many images).

**Figure 3**

*Size of Image Sets*

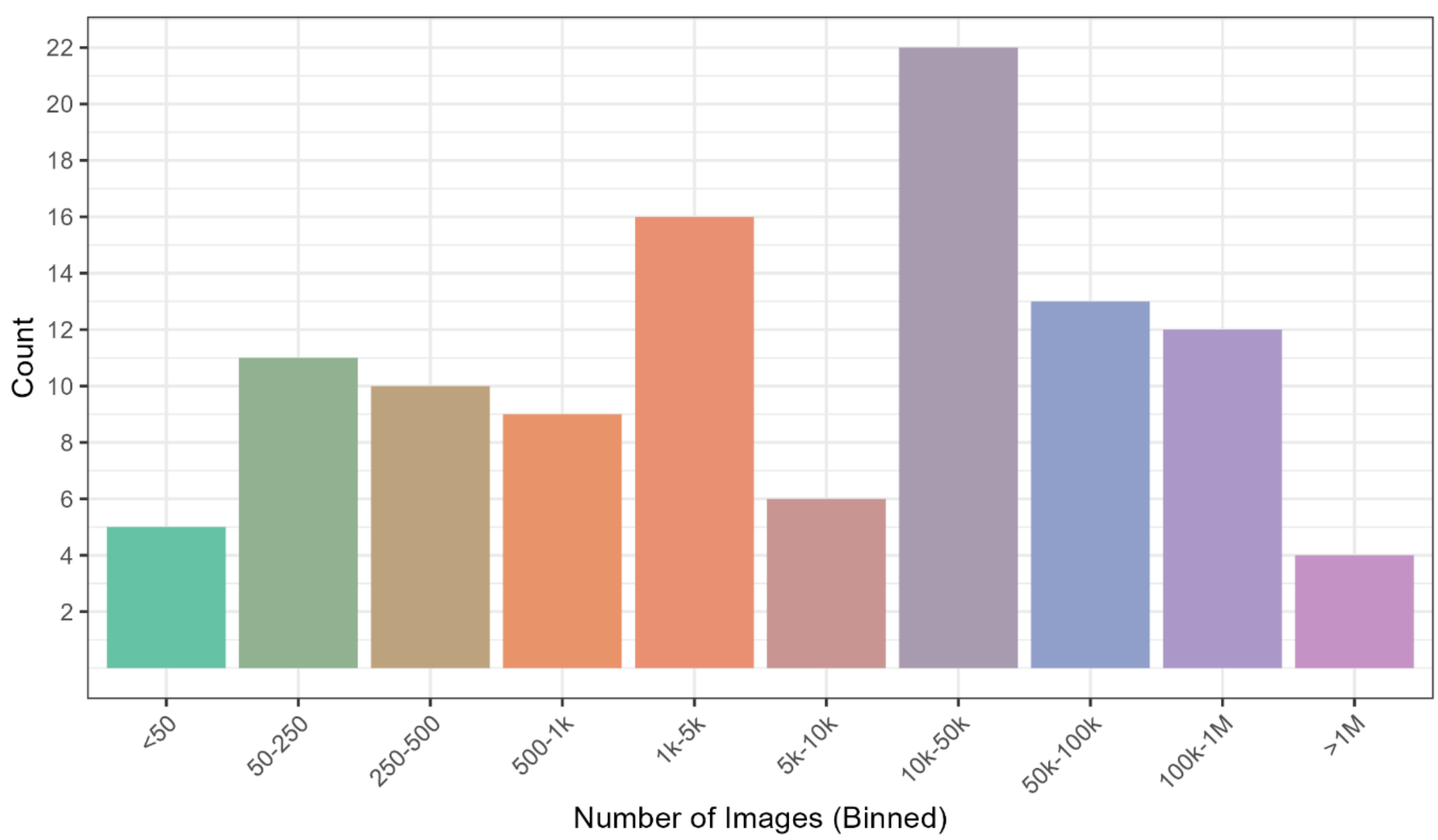

*Note.* Binned Size of Image Sets in DODA. On the x-axis the number of images per bin are plotted, on the y-axis the count of datasets within that bin.

## Number of Classes/Semantic Categories

'Classes' or 'Semantic Categories', such as art genres or art movements, can provide useful additional annotations in an image set, but not all datasets contain this kind of information. Some examples of datasets that do include these additional annotations are, for example, the JenAesthetics Dataset (Amirshahi et al., 2015) which includes annotations for the genre of each image (e.g., abstract, landscape, still life, portrait, nude, urban scene etc.), and the ArtBench-10 Dataset (Liao et al., 2022) which is annotated for 10 distinct art movements (Art Nouveau, Baroque, Expressionism, Impressionism, Post Impressionism, Realism, Renaissance, Romanticism, Surrealism, and Ukiyo-e). The Waterloo IAA image set features 5 genres: Animals, Architectures/City Scenes, Humans, Natural Scenes, and Still Object (Liu & Wang, 2017).

**Image Category and Content**

Within the field of aesthetics, the object of aesthetic evaluation can differ vastly. Some datasets focus exclusively on artworks in the form of paintings, others consist only of photographs, artificially generated images, stimuli created for an explicit research question, or a mixture.

The 'Style of Images' column lists the predominant category of images within the image set. Possible image categories include paintings, photographs, artificially created images, artificial stimuli, and vectors. We introduced the style 'Artificial Stimuli' to denote datasets of artificially generated images that represent geometric patterns (Chipman, 1977; Chipman & Mendelson, 1979; Friedenberg, 2019; Gartus & Leder, 2013; Güçlütürk et al., 2016; Jacobsen & Höfel, 2002; Nath et al., 2024; Palmer & Griscom, 2013; Wilson & Chatterjee, 2005), fractals (Bies et al., 2016; Spehar et al., 2016), or polygons (Bertamini et al., 2016; Clemente et al., 2023; Eisenman & Grove, 1972; Z. Sun & Firestone, 2022) or line drawings (Bertamini & Sinico, 2021) were used. The sub-category "Vectors" currently only applies to a single set, the "WikiArt Vectors" (Srinivasa Desikan et al., 2022) which contains color and style vectors for a subset of the online art collection WikiArt (*WikiArt.Org - Visual Art Encyclopedia*, n.d.). Some image sets contain a mixture of image styles, sometimes not only including paintings and photographs but also images of sculptures, textiles, murals etc. (Estrada Gonzalez et al., 2025; Liao et al., 2022)

The 'Overall Content' column aims to refine this initial categorization by giving more details when available, for example the cultural provenance of paintings, or content types of the categories mentioned above (e.g., landscapes, nature...).

**Number of Annotators/Participants/Captions per Image**

People differ in their aesthetic judgements (Palmer et al., 2013; Spehar et al., 2016; Van Geert & Wagemans, 2020). Therefore, the number of participants in a rating experiment is highly relevant for the inferences one would like to make about the aesthetic quality of an image. Because large image sets are often partitioned into smaller sets that are used for the testing of individual participants, the total number of participants involved in data collection is not always equal to the number of ratings per image. We therefore include two columns with this type of information. To be conservative, in case of multiple experiments across the dataset, we report the smallest sample size to be conservative. Unfortunately, some authors do not report any of these metrics, in which case the column is left empty. Currently, D-Visa (Kim et al., 2023) has the smallest number of annotators in DODA, with 3 annotators for 2,782 art images. The Minimum Semantic Content (MSC) image dataset (Parraga et al., 2024) contains ratings from more than 10,000 raters in total, and 100 ratings per image, for 10,426 images, making it the dataset with the most annotators currently listed in DODA. Notably, a subset of the Vienna Art Picture System (VAPS) (Fekete et al., 2022) features 120 raters per image. Some datasets are created by sourcing image captions from photo critique platforms (Chang et al., 2017, 2017; Ghosal et al., 2019; Jin et al., 2019; Vera Nieto et al., 2022; Zhong et al., 2022; Zhou et al., 2022), which can include tens of thousands of captions (but often of lower quality or relevance).

**Figure 4**

*Minimum Data Available per Image*

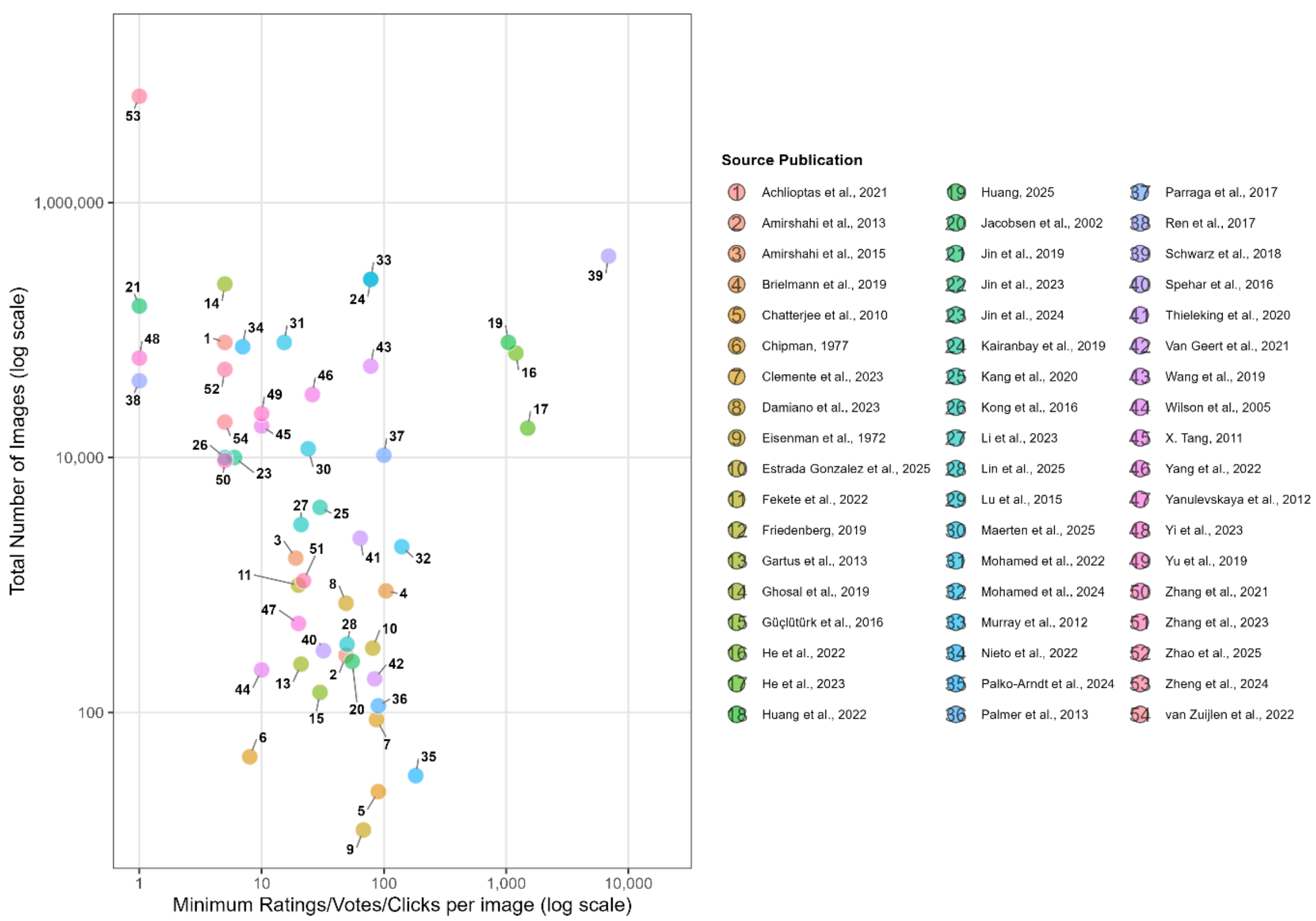

*Note.* Fifty four exemplary DODA datasets with their minimum number of ratings/votes or clicks available to illustrate the variability in rater and image numbers. The x-axis plots the minimum data available per image (e.g., minimum number or ratings, minimum number of social media likes etc.) and the y-axis the number of images in the dataset. The axes are displayed on a logarithmic scale to accommodate the range in dataset sizes and available data. The legend is ordered according to Dataset titles.

## Availability, Format, Description and Category of Data

Under the 'Data Available' column, we try to provide an overview for all aesthetic determinants or outcomes that an image set is annotated for, as well as semantic annotations. Next to annotations for overall aesthetic outcomes (e.g., scores for liking, pleasure, beauty) and semantic or style categories (e.g., art style, artist, genre), some datasets also feature annotations for other image properties. The AMD-A dataset (Jin et al., 2022), for example, provides labels for color, light and composition. The VAPS dataset (Fekete et al., 2022) provides ratings on emotional valence, emotional arousal, visual complexity and familiarity in addition to ratings on liking. The Neatly Organized Things dataset (Van Geert & Wagemans, 2021) provides ratings for the aesthetic determinants of order and complexity, as well as soothingness and fascination as outcome measures. The JenAesthetics dataset (Amirshahi et al., 2015) includes five different aesthetic annotations: aesthetic quality, beauty, liking of color, liking of content and liking of composition. In addition to the annotations provided by the authors of each dataset, DODA also provides the quantitative image properties calculated by the Aesthetics Toolbox whenever images were available and image set size is equal to or below 17,000 images. This cutoff value has pragmatic reasons: it allows us to calculate image properties for as many DODA entries as possible, whilst calculations remain feasible despite the very large size of some of the datasets. With the Aesthetic Toolbox, interested researchers can easily calculate the image properties for all datasets of their interest, should they not be provided by us already.

The ‘Data Format’ section summarizes the format of the available data in one word. Possible options here are ratings, labels, choice, counts, semantic annotations, and vectors. In case of multiple annotation formats, if the image sets contain a rating, rating will be the format mentioned in this section. ‘Ratings’ refer to image sets which were annotated using a rating task (e.g., using Likert-scales, visual analogue scales etc.). ‘Labels’ is used when the participants were asked to assign labels to the images. An example for label annotation is the study by Kim et al. (2023), in which participants were asked to assign emotion labels to paintings. In case of a preference task, as used by Sun et al. (2017) or Huang et al. (2025), we use the shorthand ‘Choice’. Whenever images were annotated through online contests (Yi et al., 2023) or social media likes (Segalin et al., 2017), we refer to 'Counts'. 'Semantic annotations' encompasses information that is potentially provided by participants of an experiment but can also be any other information related to the images (e.g., in the case of paintings, the painter, genre, art movement etc.). 'Vectors' are provided alongside the images of the study by Desikan et al. (2022).

Often, when attempting aesthetic prediction, or making statements about the aesthetic appreciation of images, researchers report the mean of aesthetic ratings. Recent work has again shown the limitations of this statistical measure (Pombo et al., 2024), as the mean fails to account for the variability in aesthetic judgements. To counteract this deficit, Rubio and colleagues (2022) suggest using the Aesthetic Ranking Value (ARV) or Weighted ARV (WARV), two metrics that are more evenly distributed than the mean and can be used as output in regression or classification tasks or in deep learning models. However, this approach will also fail to capture the variation to its fullest extent. If one is interested in generalized image aesthetic assessment (GIAA), the mean might oftentimes suffice, but as soon as individual preferences or the prediction of a personalized image aesthetic assessment (PIAA) are of interest, the mean is not sufficient (Chen et al., 2025; Ren et al., 2017).

Unfortunately, many datasets are published not with the raw data, but the means only. This makes it impossible for researchers who wish to reuse a dataset to assess the variability and reliability of the ratings. We therefore urge researchers to share their full dataset to facilitate reuse.

Similar to how 'Data Format' provides a way to filter for 'Data Available', we created the column 'Data Category' to enable users to search for the kind of annotations they are looking for more broadly. We aggregated the more detailed information into four distinct thematic categories: Aesthetic Score & Preference, Visual & Compositional Annotations, Emotions & Affect, and Content, Semantics & Captions. We used hierarchical classification, meaning whenever an overall Aesthetic core is provided (encompassing all aesthetic outcomes mentioned in the Introduction above), or a preference judgement was made, this column will reflect this fact, even if other annotations are also available. If there is no such annotation, move to Visual & Compositional Annotations and so on.

**Figure 5**

*Grouped Annotation Categories Across Datasets*

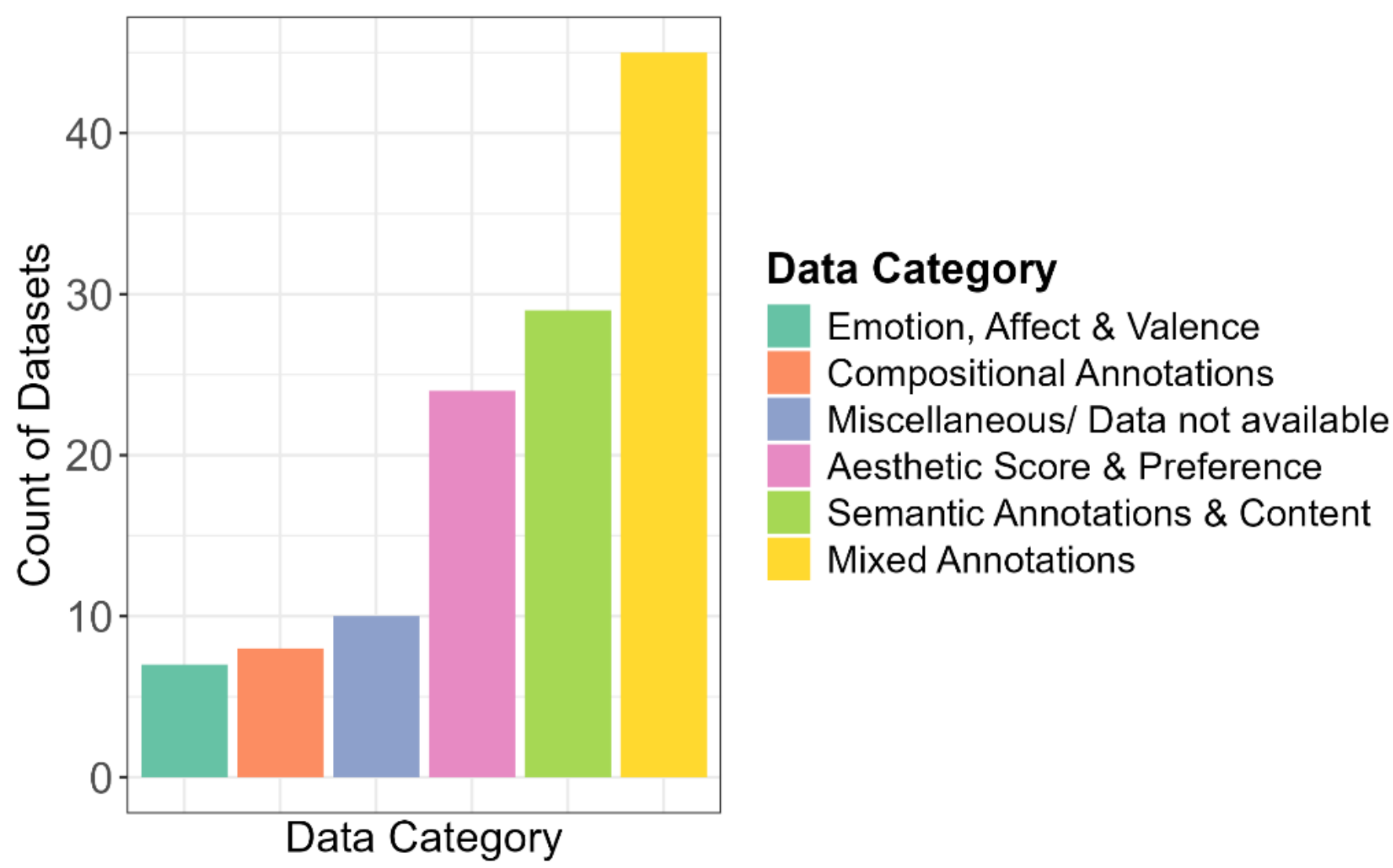

*Note.* Frequency of grouped annotation categories across datasets. Free-text metadata entries describing available dataset annotations ('Data Available' Column) were combined into four overarching themes. The y-axis indicates the number of datasets containing at least one variable belonging to the respective category.

## Scale of Annotations and Their Meaning

To interpret any annotation of the images, and to successfully compare scores across image sets, it is essential to specify annotations and their generation procedure. DODA therefore includes a column that contains the rating scale used, as well as a second column explaining how data was collected more generally (e.g., the experimental setup). The scales could include, for example, Likert scales of different length (Chatterjee et al., 2010; Damiano et al., 2023; Fekete et al., 2022), bi-polar scales (Chang et al., 2017; Vartanian et al., 2013; J. Wang et al., 2024; Wilson & Chatterjee, 2005), Visual Analogue Scales (VAS) (Maerten et al., 2025; Nath et al., 2024) or perhaps a forced-choice preference task. In some datasets, the scores are obtained by counting the "likes" that the images have gotten on a social site (Schwarz et al., 2018; Segalin et al., 2017; Yi et al., 2023). Whenever possible we also try to share a short statement on how the data was collected in the 'How are scores obtained?' column.

**Image Source and Source Category**

Oftentimes, digital images are obtained from datasets available in the Web or from other datasets. For paintings, WikiArt[2], a non-profit digital art collection, is a commonly used source. Photographs are often extracted from online photography challenges like DPChallenge[3] or from photo-sharing sites like Flickr[4]. The source of the datasets is important to consider for two reasons: on the one hand, the fact that these collections are preselected with a certain goal (e.g., submitting an especially stunning photo to gain followers or win the challenge on a specific theme) can skew the distribution of aesthetic values within the image set. Thus, a high score in a "Black & White" photography challenge, might not be directly comparable in aesthetic value to a high score in a "Pet Photography" challenge. On the other hand, with the rapid increase in published datasets it is harder to keep track of potential overlap between them, especially when the datasets are large. For example, EVA (Kang et al., 2020) is a subset of AVA (Murray et al., 2012), which in turn is created from DPChallenge images. DODA aims to provide the earliest traceable image source to help users avoid unwanted overlap. For easier filtering, we added the 'Source Category' column where we tried to summarize the possible sources. The 'Online Art & Photography Communities' category includes crowdsourced platforms and contest sites (e.g., Flickr, DPChallenge, Boldbrush[5]); 'Museums, Books & Curated Collections' comprises curated digitizations; 'Custom/Experimental Stimuli' refers to images specifically designed or captured by the study authors and their collaborators (using algorithms or working with art historians or designers, for example); 'AI Generated' includes content from generative models like Stable Diffusion; 'Other Non-Aesthetic/Miscellaneous' includes images from datasets that were previously used for non-aesthetics research and 'General Web Search' denotes datasets compiled via broad search engine queries. Image sets with multiple possible labels were classified as 'Other/Mixed'.

---

[2] https://www.wikiart.org/
[3] https://www.dpchallenge.com/
[4] https://www.flickr.com/

We would like to note that some museums provide high-quality open-source images from their collections. We find this practice commendable and have compiled a list of museums that do so (see Table 2 of the Supplementary Materials).

[5] https://faso.com/boldbrush/

**Figure 6**

*Classification of image source*

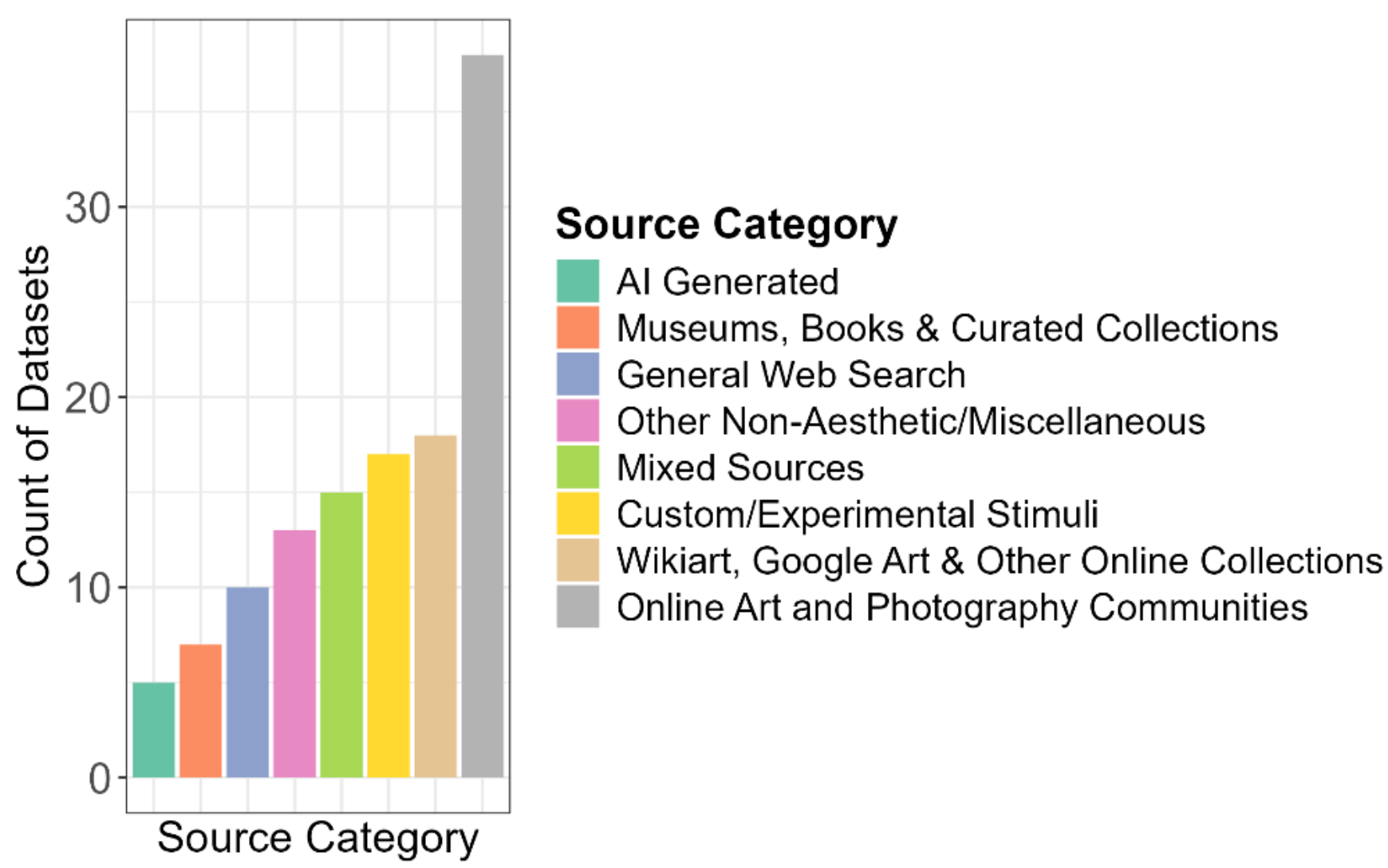

*Note.* Image sources were categorized into eight primary groups based on their origin for the 'Source Category' column of DODA.

**Field of Most Common Use and Task Specification**

DODA also includes a 'Field of Most Common Use' column, in which we assess whether an image set has been used predominantly in empirical or computational aesthetics. This is intended as a filtering tool, in case users from a specific subdiscipline wish to explore image sets that might be unfamiliar to them. Additionally, we also include a column to filter for the task the dataset was likely created for. Datasets with annotations are more likely suited for classification tasks (e.g., identifying objects within images, creators or styles of artworks etc.), whereas datasets with scores are more likely created to aid in the prediction of aesthetic preference or understanding of aesthetic scores.

**Resolution, Image Quality and Image Fidelity**

Image Quality Assessment is not the same as Image Aesthetic Assessment. Digital images can undergo many types of distortions during their acquisition, processing, compression, storage, transmission, or reproduction, all of which can reduce their visual quality (Z. Wang et al., 2004). Therefore, Image Quality Assessment is a prominent subfield of computer vision that explores how to assess, maintain, and improve the quality of digital images (Athar & Wang, 2019). Broadly, Image Quality Assessment can be achieved through two kinds of measures: subjective and objective. Since most images are meant for human viewing, the most relevant way to measure visual quality is through the subjective evaluation by people (Z. Wang & Bovik, 2006). However, this approach is often impractical because it is time-consuming, costly, and inconvenient. Therefore, objective image quality assessment research aims to create quantitative methods that can automatically predict how humans perceive image quality (Z. Wang & Bovik, 2006). This complex task involves many parameters such as contrast, focus, illumination, changes in color spectrum and more. Some of these parameters overlap with the QIPs mentioned in the introduction. QIPs, such as contrast, lightness entropy, complexity, and Fourier slope, can be easily calculated using the Aesthetics Toolbox via an intuitive web application or Python script (Redies et al., 2025). For the user's convenience we provide pre-calculated QIPs for all datasets found on DODA.

Subjective and objective image quality measures may be closely related, but can also differ in important aspects. For example, some distortions measured objectively can have different perceptual thresholds, i.e., people may differ in their sensitivity to certain image distortions (Pin & Amirshahi, 2025). Furthermore, they might perceive a distortion caused by blur as more negatively affecting quality than a distortion caused by high luminance levels.

We can think of image quality as a sort of prerequisite for aesthetic quality, especially in visual aesthetics, where it is of utmost importance to have the digital reproduction resemble the original painting as closely as possible. Recent work has shown that participants prefer images of higher fidelity when provided different versions of the same image (Koßmann et al., 2025). Unfortunately, many artworks exist online in non-curated forms (e.g., cropped, blurred, colour-shifted, or digitally altered otherwise). When image fidelity is not guaranteed, the generalizability of research results from the reproductions to the original artworks may be drawn into question (De Winter & Koßmann, 2026). Of course, in case of photography, image distortions can be a stylistic devise, as a photographer might choose to play with blur, highlights or purposefully create other visual effects.

In summary, image quality can play an important role when evaluating the aesthetics of images. Therefore, assessing the perceptual quality of reproductions should be part of all stimulus selection. For this assessment, one might use the algorithmic quality assessment approach described above, which is especially useful when many images are needed, as an alternative to costly ratings by humans (Athar & Wang, 2019; Ding et al., 2021), though it must be noted that computational quality assessment might diverge occasionally from subjective human quality assessment.

To provide users with a first, simple and popular measure of image quality, DODA provides the average resolution across all images for each dataset, wherever possible.

**Heterogeneity Score**

The heterogeneity of datasets significantly affects various types of applications, especially the performance of statistical prediction models. Learning and accurately predicting labels is easier with more homogeneous datasets, while high levels of heterogeneity present greater challenges for prediction accuracy.

Dataset heterogeneity can be quantified through various approaches, such as using image labels, content, creation times and locations, or feature maps from neural networks. In this work, we use an approach based on the quantitative image properties (QIPs) which we provide in DODA alongside most datasets. Specifically, we compute the average Shannon entropy by aggregating the Shannon entropy values of individual QIPs within each dataset as follows:

Let a dataset $D$ contain $N$ images. For a given QIP $q$, each image produces a discrete value $x_i(q)$. We define the empirical frequency of each observed value $v$ as:

$$p_q(v) = \frac{1}{N}\sum_{i=1}^{N} 1\left(x_i^{(q)} = v\right) \quad (1)$$

where **1**(·) is the indicator function. For each QIP $q$, heterogeneity is measured using normalized Shannon entropy:

$$H_q(D) = -\frac{1}{log(K_q)}\sum_{v \in \mathcal{V}_q} p_q(v) log p_q(v) \quad (2)$$

where $V_q$ is the set of all possible discrete values of QIP $q$, $K_q = |Vq|$ is the maximum possible number of bins, and the normalization by $\log(K_q)$ ensures $H_f \in [0, 1]$. The Normalization ensures that all QIPs have the same weight on the heterogeneity score. The final scalar heterogeneity score for each dataset $D$ with $M$ QIPs is the arithmetic mean over all QIP entropies:

$$H_{mean}(D) = \frac{1}{M}\sum_{q=1}^{M} H_q(D) \quad (3)$$

This score reflects the variability of low-level image statistics within a dataset: higher values indicate greater diversity in image properties (i.e., higher heterogeneity), whereas lower values indicate more uniform datasets. Importantly, this measure is agnostic to semantic content and instead captures structural and statistical variability in the images. As such, it provides a complementary perspective on dataset diversity that is particularly relevant for computational modeling, where variability in input features can strongly influence model generalization and performance.

The subset of quantitative image properties (QIPs) used for the computation of the heterogeneity score was selected based on both methodological and conceptual considerations. First, all chosen QIPs exhibit a fixed and bound range of possible values. This property is essential for the reliable estimation of Shannon entropy, which requires well-defined probability distributions. Variables with unbounded or highly skewed ranges (e.g., extending from 0 to +∞) can lead to unstable binning, sparsity issues, and reduced comparability across datasets. By restricting the analysis to QIPs with finite and interpretable value ranges, we ensure that entropy estimates are numerically stable and comparable across datasets.

Second, the selected QIPs were chosen to span multiple, complementary aspects of image structure and appearance. Specifically, they cover (i) low-level intensity variation (RMS contrast), (ii) color information in both RGB and perceptually motivated color spaces (mean R, G, B, and L channel values), and (iii) spatial and compositional properties (mirror symmetry, balance, homogeneity, and DCM distance). This combination ensures that the heterogeneity score captures variability across different dimensions of visual information, rather than being biased toward a single class of image features.

Taken together, this selection provides a compact yet diverse representation of image statistics, enabling a robust and interpretable quantification of dataset heterogeneity.

**Figure 7**

*Comparison of Dataset Heterogeneity via Mean Normalized Entropy*

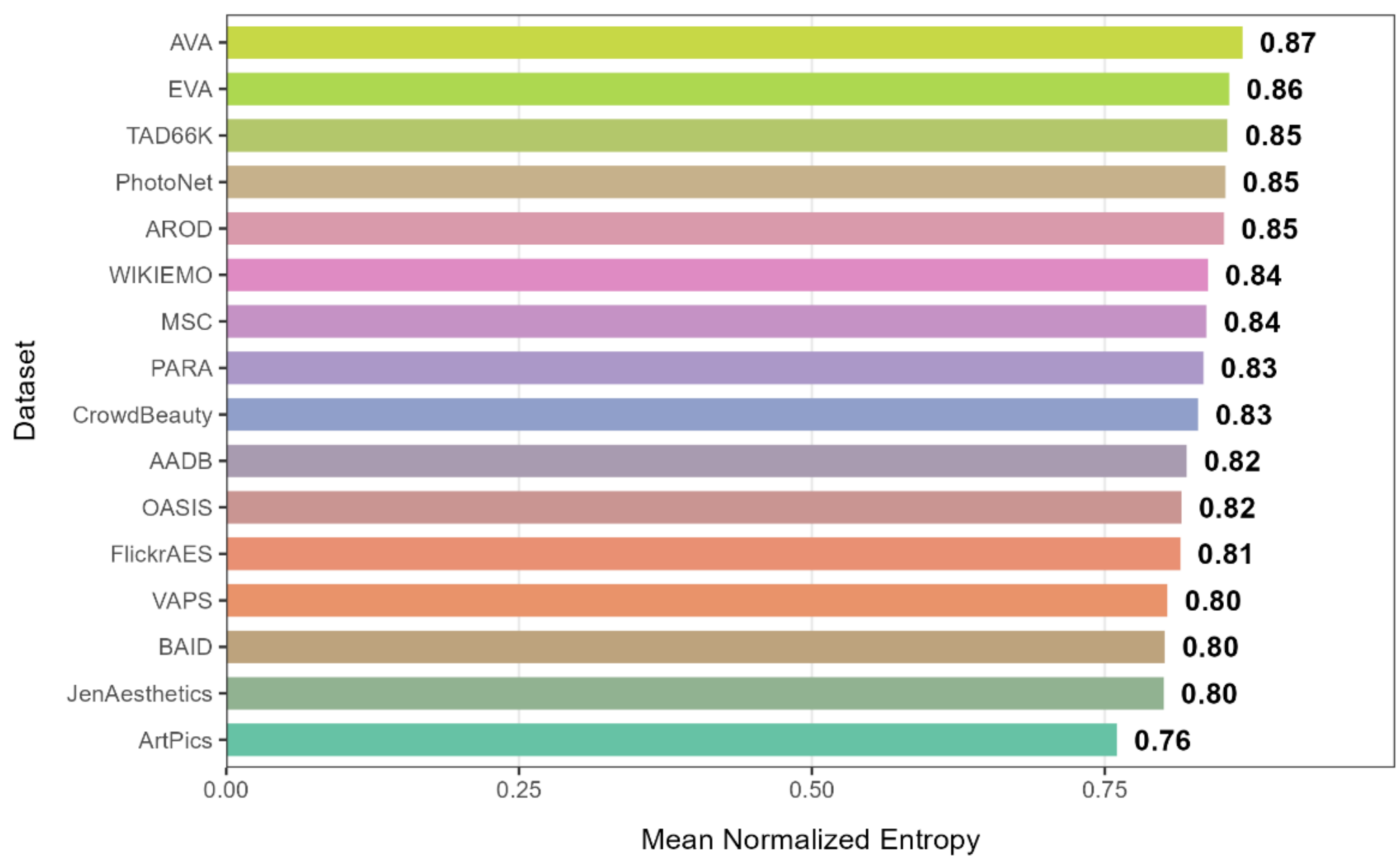

*Note*. Datasets are ordered by ascending mean entropy values. Values next to bars indicate the exact mean normalized entropy score for each dataset.

These rankings are consistent with qualitative expectations. ArtPics is the least heterogeneous dataset, as it consists of style-transferred images. JenAesthetics, BAID, and VAPS are composed of fine-art imagery and therefore exhibit relatively homogeneous visual characteristics. In contrast, FlickrAES, AADB, and CrowdBeauty share a common source in Flickr, which contributes to their comparable levels of variability. EVA is a subset of AVA and thus inherits its structural and distributional properties. Finally, datasets derived from photographic challenges, such as AVA, EVA, and PhotoNet, exhibit the highest degree of heterogeneity, which is expected given the broad diversity of themes, contexts, and photographic tasks represented in these challenges.

**Link to the Dataset and Image Copyright Restrictions**

Artworks and images are subject to copyright laws. The goal of copyright protection is to guarantee that authors receive fair compensation for producing and distributing their original works. EU-level harmonization and international treaties have aligned copyright traditions across member states. However, important discrepancies persist (Geiger & Schönherr, 2017). In the most basic terms, any photograph taken by a person is automatically protected by copyright. However, most states require a threshold of originality of the work to be eligible for copyright protection (Geiger & Schönherr, 2017).

The Berne Convention established that copyright protection persists for at least 50 years after an author's death (World Intellectual Property Organization, n.d.). The European Union extends the duration of copyright protection to encompass 70 years after the author has passed. In contrast, the duration of copyright protection in the United States does not depend on the author's death but on the year the work was made public (for works published before 1978) (The European Union Intellectual Property Office, 2025). After copyright protection expires, the work becomes part of the public domain and can be used freely. In addition, copyrighted works can be used for research purposes in some countries, for example in the United Kingdom (CDPA, 1988) and in Germany (UrhG, 1965; and later versions).

To be mindful of copyright, and because of currently limited storage space, we cannot provide a direct download link for the images in DODA. Instead, we provide a link to the storage of the original publications, where users can follow download instructions. Obviously, we cannot guarantee that all these links remain 'live', but we are willing to provide updates or corrections when we find out or are informed about it.

## Discussion

Our aim was to create a tool for researchers to browse datasets in the fields of empirical and computational aesthetics and to facilitate exchange and collaboration between researchers within and beyond these fields, in order to advance our understanding of aesthetic appreciation. To this end, we introduce the Database of Datasets for Aesthetics (DODA)[6], an intuitive online tool which also allows to filter the datasets. Moreover, we provide precomputed quantitative image properties, homogeneity values, and image resolution for the datasets.

Working with the same datasets provides the opportunity to directly and indirectly collaborate across methodologies within the field of aesthetics. Furthermore, we hope that DODA will help young researchers get started with stimulus selection and allow experienced researchers to see opportunities for more ecological research practices at a glance. DODA is part of the Aesthetics Toolbox, which encompasses an easy way to calculate quantitative image properties, and other useful tools, such as image resizing. Taken together, these tools will hopefully make aesthetics research on digital images more widely accessible in the research community.

DODA will be most useful if it keeps growing with the state-of-the-art. We therefore invite researchers to contact us to include their datasets for aesthetics research at aestheticstoolbox@gmail.com, ideally with all of the essential information to be put in the appropriate columns of DODA. Submissions will be added to DODA in a

[6] available online under https://aesthetics-toolbox.streamlit.app/DODA

timely manner. We will acknowledge the contributors by name on a dedicated page of the online presentation of the Aesthetics Toolbox.

**Acknowledgements**

The authors would like to thank Nicole Ruta for designing and creating the DODA logo and Fatemeh Behrad, Anne-Sofie Maerten, Eline Van Geert, and Li-Wei Chen for enlightening discussions about datasets used in aesthetics research and Gonzalo Muradás Odriozola for technical support. We also thank Michiel Willems for discussions about Online Museum Collections. Thanks also to our student researcher Rehaab Ziani for additional literature research. Finally, we would like to thank Nikolai Huckle, who provided a great overview of art datasets on GitHub georgeblck/art-datasets, which allowed us to include 25 additional image sets in our project.

We acknowledge the use of Gemini (https://gemini.google.com/app) as a coding tool to debug R Code that was used to create the plots in this paper and to assist in writing the R script used for dataset and image source categorization via string detection. We verified the results and retain full responsibility for them.

This work was, in part, funded by the European Union (ERC AdG, GRAPPA, 101053925, awarded to Johan Wagemans). Views and opinions expressed are, however, those of the authors only and do not necessarily reflect those of the European Union or the European Research Council Executive Agency. Neither the European Union nor the granting authority can be held responsible for them.

## References

Act on Copyright and Related Rights (Urheberrechtsgesetz) (Germany), Pub. L. No. BGBl. I S. 1273, UrhG (1965). https://www.gesetze-im-internet.de/englisch_urhg/englisch_urhg.html

Amirshahi, S. A., Hayn-Leichsenring, G. U., Denzler, J., & Redies, C. (2015). JenAesthetics Subjective Dataset: Analyzing paintings by subjective scores. *Lecture Notes in Computer Science*, *8925*, 3–19. https://doi.org/10.1007/978-3-319-16178-5_1

Arnheim, R. (1954). *Art and visual perception: A psychology of the creative eye*. University of California Press.

Athar, S., & Wang, Z. (2019). A comprehensive performance evaluation of image quality assessment algorithms. *IEEE Access*, *7*, 140030–140070. https://doi.org/10.1109/ACCESS.2019.2943319

Augustin, M. D., Wagemans, J., & Carbon, C.-C. (2012). All is beautiful? Generality vs. specificity of word usage in visual aesthetics. *Acta Psychologica*, *139*(1), 187–201. https://doi.org/10.1016/j.actpsy.2011.10.004

Bartho, R., Thoemmes, K., & Redies, C. (2023). *Predicting beauty, liking, and aesthetic quality: A comparative analysis of image databases for visual aesthetics research*. arXiv.https://arxiv.org/abs/2307.00984v1

Bertamini, M., Palumbo, L., Gheorghes, T. N., & Galatsidas, M. (2016). Do observers like curvature or do they dislike angularity? *British Journal of Psychology*, *107*(1), 154–178. https://doi.org/10.1111/BJOP.12132

Bertamini, M., & Sinico, M. (2021). A study of objects with smooth or sharp features created as line drawings by individuals trained in design. *Empirical Studies of the Arts*, *39*(1), 61–77. https://doi.org/10.1177/0276237419897048

Bies, A. J., Blanc-Goldhammer, D. R., Boydston, C. R., Taylor, R. P., & Sereno, M. E. (2016). Aesthetic responses to exact fractals driven by physical complexity. *Frontiers in Human Neuroscience*, *10*. https://doi.org/10.3389/fnhum.2016.00210

Brachmann, A., & Redies, C. (2017). Computational and experimental approaches to visual aesthetics. *Frontiers in Computational Neuroscience*, *11*. https://doi.org/10.3389/fncom.2017.00102

Brielmann, A. A., & Pelli, D. G. (2018). Aesthetics. *Current Biology*, *28*(16), R859–R863. https://doi.org/10.1016/J.CUB.2018.06.004

Brielmann, A. A., & Pelli, D. G. (2019). Intense beauty requires intense pleasure. *Frontiers in Psychology*, *10*. https://doi.org/10.3389/fpsyg.2019.02420

Chamberlain, R. (2022). The interplay of objective and subjective factors in empirical aesthetics. In B. Ionescu, W. A. Bainbridge, & N. Murray (Eds.), *Human Perception of Visual Information: Psychological and Computational Perspectives* (pp. 115–132). Springer International Publishing. https://doi.org/10.1007/978-3-030-81465-6_5

Chang, K.-Y., Lu, K.-H., & Chen, C.-S. (2017). Aesthetic critiques generation for photos. *Proceedings of the IEEE International Conference on Computer Vision*, *2017*, 3534–3543. https://doi.org/10.1109/ICCV.2017.380

Chatterjee, A., & Vartanian, O. (2016). Neuroscience of aesthetics. *Annals of the New York Academy of Sciences*, *1369(1),* 172–194. https://doi.org/10.1111/nyas.13035

Chatterjee, A., Widick, P., Sternschein, R., Smith, W., & Bromberger, B. (2010). The assessment of art attributes. *Empirical Studies of the Arts*, *28*(2), 207–222. https://doi.org/10.2190/EM.28.2.f

Chen, L.-W., Strafforello, O., Maerten, A.-S., Tuytelaars, T., & Wagemans, J. (2025). On the role of individual differences in current approaches to computational image aesthetics. *Proceedings of the British Machine Vision Conference*, *2025*, Article 680. https://bmva-archive.org.uk/bmvc/2025/assets/papers/Paper_680/paper.pdf

Chipman, S. F. (1977). Complexity and structure in visual patterns. *Journal of Experimental Psychology: General*, *106*(3), 269–301. https://doi.org/10.1037/0096-3445.106.3.269

Chipman, S. F., & Mendelson, M. J. (1979). Influence of six types of visual structure on complexity judgments in children and adults. *Journal of Experimental Psychology: Human Perception and Performance*, *5*(2), 365–378. https://doi.org/10.1037/0096-1523.5.2.365

Clemente, A., Penacchio, O., Vila-Vidal, M., Pepperell, R., & Ruta, N. (2025). Explaining the *curvature effect*: Perceptual and hedonic evaluations of visual contour. *Psychology of Aesthetics, Creativity, and the Arts, 19*(5), 933–952. https://doi.org/10.1037/aca0000561

Copyright, Designs and Patents Act 1988, c. 48 Public General Acts § 29 (1988). https://www.legislation.gov.uk/ukpga/1988/48/section/29

Cupchik, G. C., & Gebotys, R. J. (1990). Interest and pleasure as dimensions of aesthetic response. *Empirical Studies of the Arts*, *8*(1), 1–14. https://doi.org/10.2190/L789-TPPY-BD2Q-T7TW

Damiano, C., Wilder, J., Zhou, E. Y., Walther, D. B., & Wagemans, J. (2023). The role of local and global symmetry in pleasure, interest, and complexity judgments of natural scenes. *Psychology of Aesthetics, Creativity, and the Arts*, *17*(3), 322–337. https://doi.org/10.1037/aca0000398

De Winter, S., & Koßmann, L. (2026, September 14–18). *A conservation approach to digital image datasets: Safeguarding cultural integrity in the digital age* [Paper presentation]. 21st International Council of Museums Committee for Conservation (ICOM-CC) Triennial Conference, Oslo, Norway.

Ding, K., Ma, K., Wang, S., & Simoncelli, E. P. (2021). Comparison of full-reference image quality models for optimization of image processing systems. *International Journal of Computer Vision*, *129*, 1258–1281. https://doi.org/10.1007/s11263-020-01419-7

Eisenman, R., & Grove, M. S. (1972). Self-ratings of creativity, semantic differential ratings, and preferences for polygons varying in complexity, simplicity, and symmetry. *The Journal of Psychology: Interdisciplinary and Applied*, *81*(1), 63–67. https://doi.org/10.1080/00223980.1972.9923789

Estrada Gonzalez, V., Bobrow, I., Cardillo, E. R., Kim, O., Meletaki, V., & Chatterjee, A. (2025). A normed art database that incorporates diverse cultures and genres:

The Penn Center for Neuroaesthetics artwork repository. *Psychology of Aesthetics, Creativity, and the Arts*. https://doi.org/10.1037/aca0000755

Fekete, A., Pelowski, M., Specker, E., Brieber, D., Rosenberg, R., & Leder, H. (2022). The Vienna Art Picture System (VAPS): A data set of 999 paintings and subjective ratings for art and aesthetics research. *Psychology of Aesthetics, Creativity, and the Arts*, *17*(5), 660–671. https://doi.org/10.1037/ACA0000460

Friedenberg, J. (2019). Beauty in the eye of the beholder: Individual differences in preference for randomized visual patterns. *Experimental Psychology*, *66*(2), 112–125. https://doi.org/10.1027/1618-3169/a000432

Gartus, A., & Leder, H. (2013). The small step toward asymmetry: Aesthetic judgment of broken symmetries. *I-Perception*, *4*(5), 361–364. https://doi.org/10.1068/i0588sas

Geiger, C., & Schönherr, F. (2017). *Consumers' Frequently Asked Questions (FAQS) on Copyright* (pp. 1–96) [Summary Report]. The European Union Intellectual Property Office. https://euipo.europa.eu/tunnel-web/secure/webdav/guest/document_library/observatory/documents/div/FAQs%20on%20Copyright,%20Summary%20Report%20January%202017.pdf

Geller, H. A., Bartho, R., Thömmes, K., & Redies, C. (2022). Statistical image properties predict aesthetic ratings in abstract paintings created by neural style transfer. *Frontiers in Neuroscience*, *16*. https://doi.org/10.3389/fnins.2022.999720

Ghosal, K., Rana, A., & Smolic, A. (2019). Aesthetic Image captioning from weakly-labelled photographs. *Proceedings - 2019 International Conference on Computer Vision Workshop, ICCVW 2019*, 4550–4560. https://doi.org/10.1109/ICCVW.2019.00556

Graf, L. K. M., & Landwehr, J. R. (2015). A dual-process perspective on fluency-based aesthetics: The pleasure-interest model of aesthetic liking. *Personality and Social Psychology Review*, *19*(4), 395–410. https://doi.org/10.1177/1088868315574978

Graf, L. K. M., & Landwehr, J. R. (2017). Aesthetic pleasure versus aesthetic interest: The two routes to aesthetic liking. *Frontiers in Psychology*, *8*(JAN). https://doi.org/10.3389/fpsyg.2017.00015

Güçlütürk, Y., Jacobs, R. H. A. H., & Lier, R. van. (2016). Liking versus complexity: Decomposing the inverted U-curve. *Frontiers in Human Neuroscience*, *10*. https://doi.org/10.3389/fnhum.2016.00112

Hook, D., & Glaveanu, V. P. (2013). Image analysis: An interactive approach to compositional elements. *Qualitative Research in Psychology*, *10*(4), 355–368. https://doi.org/10.1080/14780887.2012.674175

Huang, L. (2025). Aesthetic preferences among spatial patterns: Large-scale experiment, comprehensive exploration, and a three-component regularity-based model. *Psychology of Aesthetics, Creativity, and the Arts*. https://doi.org/10.1037/aca0000774

Huckle, N. [georgeblck]. (2020a). *Art-datasets* [Computer software]. GitHub. Retrieved June 23, 2026, from https://github.com/georgeblck/art-datasets

Jacobsen, T. (2006). Bridging the Arts and Sciences: A framework for the psychology of aesthetics. *Leonardo*, *39*(2), 155–162. https://doi.org/10.1162/LEON.2006.39.2.155

Jacobsen, T., & Höfel, L. (2002). Aesthetic judgments of novel graphic patterns: Analyses of individual judgments. *Perceptual and Motor Skills*, *95*(3 PART 1), 755–766. https://doi.org/10.2466/PMS.2002.95.3.755

Jin, X., Wu, L., Zhao, G., Li, X., Zhang, X., Ge, S., Zou, D., Zhou, B., & Zhou, X. (2019). Aesthetic attributes assessment of images. *Proceedings of the 27th ACM International Conference on Multimedia*, 311–319. https://doi.org/10.1145/3343031.3350970

Kant, I. (1790). *Critik der Urtheilskraft*. Lagarde und Friedrich.

Kenett, Y. N., Cardillo, E. R., Christensen, A. P., & Chatterjee, A. (2023). Aesthetic emotions are affected by context: A psychometric network analysis. *Scientific Reports*, *13*(1), 20985. https://doi.org/10.1038/s41598-023-48219-w

Kim, S., An, C., Cha, J., Kim, D., & Park, E. (2023). D-ViSA: A dataset for detecting visual sentiment from art images. *2023 IEEE/CVF International Conference on Computer Vision Workshops (ICCVW)*, 3043–3051. https://doi.org/10.1109/ICCVW60793.2023.00328

Koßmann, L., De Winter, S., Woussen, J., Bossens, C., & Wagemans, J. (2025, September 11–12*).* Fidelity loss in Wiki Art images of paintings: Effects on aesthetic ratings and preferences [Talk]. *Sixteenth International Conference on The Image*, Paris, France.

Koßmann, L., Hellemans, A.-S., & Wagemans, J. (2026). Composition and spatial layout: Aesthetic determinants and outcomes in natural images and artworks. *Manuscript in preparation*.

Koßmann, L., Meulemans, M., Van Geert, E., De Winter, S., Willems, M., & Wagemans, J. (2026). Manipulating aesthetic judgements of composition through systematic variation of compositional principles. *Manuscript in preparation*.

Kurdi, B., Lozano, S., & Banaji, M. R. (2017). Introducing the Open Affective Standardized Image Set (OASIS). *Behavior Research Methods*, *49*(2), 457–470. https://doi.org/10.3758/s13428-016-0715-3

LAION Project. (2022). *LAION-Aesthetics* (Version 1) [Dataset]. GitHub. https://github.com/LAION-AI/laion-datasets/blob/main/laion-aesthetic.md

Leder, H., Belke, B., Oeberst, A., & Augustin, D. (2004). A model of aesthetic appreciation and aesthetic judgments. *British Journal of Psychology*, *95*(4), 489–508. https://doi.org/10.1348/0007126042369811

Leder, H., & Nadal, M. (2014). Ten years of a model of aesthetic appreciation and aesthetic judgments: The aesthetic episode – Developments and challenges in empirical aesthetics. *British Journal of Psychology*, *105*(4), 443–464. https://doi.org/10.1111/BJOP.12084

Liao, P., Li, X., Liu, X., & Keutzer, K. (2022). *The ArtBench Dataset: Benchmarking generative models with artworks*. https://arxiv.org/abs/2206.11404v1

Lin, Y., Op de Beeck, H., & Wagemans, J. (2025). Leuven Orthogonalized Art Data Set (LOAD): A multidimensional art image set for aesthetic appreciation research. *Psychology of Aesthetics, Creativity, and the Arts*. https://doi.org/10.1037/aca0000791

Liu, L., Zhang, X., Xing, Y., Deng, Y., Yang, Y., Zhang, Z., & Wang, G. [dieuroi]. (2023). *Awesome-Image-Aesthetic-Assessment* [Computer Software]. GitHub.

Retrieved June 23, 2026, from https://github.com/dieuroi/Awesome-Image-Aesthetic-Assessment

Liu, W., & Wang, Z. (2017). A database for perceptual evaluation of image aesthetics. *2017 IEEE International Conference on Image Processing (ICIP)*, 1317–1321. https://doi.org/10.1109/ICIP.2017.8296495

Lyssenko, N., Redies, C., & Hayn-Leichsenring, G. U. (2016). Evaluating abstract art: Relation between term usage, subjective ratings, image properties and personality traits. *Frontiers in Psychology*, *7*. https://doi.org/10.3389/fpsyg.2016.00973

*Maarten Wijntjes*. (n.d.). Maarten Wijntjes. Retrieved April 9, 2026, from https://maartenwijntjes.github.io/

Maerten, A.-S., Chen, L.-W., Winter, S. D., Bossens, C., & Wagemans, J. (2025). LAPIS: A novel dataset for personalized image aesthetic assessment. *Proceedings of the IEEE/CVF Conference on Computer Vision and Pattern Recognition (CVPR) Workshops*, 6302–6311. https://doi.org/https://doi.org/10.48550/arXiv.2504.07670

Makin, A. D. J., Helmy, M., & Bertamini, M. (2018). Visual cortex activation predicts visual preference: Evidence from Britain and Egypt. *Quarterly Journal of Experimental Psychology*, *71*(8), 1771–1780. https://doi.org/10.1080/17470218.2017.1350870

Makin, A. D. J., Wright, D., Rampone, G., Palumbo, L., Guest, M., Sheehan, R., Cleaver, H., & Bertamini, M. (2016). An electrophysiological index of perceptual goodness. *Cerebral Cortex*, *26*(12), 4416–4434. https://doi.org/10.1093/cercor/bhw255

McManus, I. C., Stöver, K., & Kim, D. (2011). Arnheim's Gestalt theory of visual balance: Examining the compositional structure of art photographs and abstract images. *I-Perception*, *2*(6), 615–647. https://doi.org/10.1068/i0445aap

Menninghaus, W., Wagner, V., Wassiliwizky, E., Schindler, I., Hanich, J., Jacobsen, T., & Koelsch, S. (2019). What are aesthetic emotions? *Psychological Review*, *126*(2), 171–195. https://doi.org/10.1037/rev0000135

Mohamed, Y., Abdelfattah, M., Alhuwaider, S., Li, F., Zhang, X., Church, K., & Elhoseiny, M. (2022a). ArtELingo: A million emotion annotations of WikiArt with emphasis on diversity over language and culture. In Y. Goldberg, Z. Kozareva, & Y. Zhang (Eds.), *Proceedings of the 2022 Conference on Empirical Methods in Natural Language Processing* (pp. 8770–8785). Association for Computational Linguistics. https://doi.org/10.18653/v1/2022.emnlp-main.600

Mohamed, Y., Li, R., Ahmad, I. S., Haydarov, K., Torr, P., Church, K., & Elhoseiny, M. (2024, November). No culture left behind: ArtELingo-28, a benchmark of WikiArt with captions in 28 languages. In Y. Al-Onaizan, M. Bansal, & Y.-N. Chen (Eds.), *Proceedings of the 2024 Conference on Empirical Methods in Natural Language Processing* (pp. 20939–20962). Association for Computational Linguistics. https://doi.org/10.18653/v1/2024.emnlp-main.1165

Nadal, M., & Vartanian, O. (2022). Empirical aesthetics: An overview. In M. Nadal & O. Vartanian (Eds.), *The Oxford Handbook of Empirical Aesthetics* (p. 0). Oxford University Press. https://doi.org/10.1093/oxfordhb/9780198824350.013.1

Nascimento, S. M. C., Marit Albers, A., & Gegenfurtner, K. R. (2021). Naturalness and aesthetics of colors – Preference for color compositions perceived as natural. *Vision Research*, *185*, 98–110. https://doi.org/10.1016/j.visres.2021.03.010

Nath, S. S., Brändle, F., Schulz, E., Dayan, P., & Brielmann, A. (2024). Relating objective complexity, subjective complexity, and beauty in binary pixel patterns. *Psychology of Aesthetics, Creativity, and the Arts*. https://doi.org/10.1037/aca0000657

Vera Nieto, D., Celona, L., & Fernandez-Labrador, C. (2022). Understanding aesthetics with language: A photo critique dataset for aesthetic assessment. *Advances in Neural Information Processing Systems*, *35*, 34148–34161. https://proceedings.neurips.cc/paper_files/paper/2022/file/dcd18e50ebca0af89187c6e35dabb584-Paper-Datasets_and_Benchmarks.pdf

Palmer, S. E., & Griscom, W. S. (2013). Accounting for taste: Individual differences in preference for harmony. *Psychonomic Bulletin & Review*, *20*(3), 453–461. https://doi.org/10.3758/s13423-012-0355-2

Palmer, S. E., Schloss, K. B., & Sammartino, J. (2013). Visual aesthetics and human preference. *Annual Review of Psychology*, *64*(Volume 64, 2013), 77–107. https://doi.org/10.1146/annurev-psych-120710-100504

Parraga, C. A., Muñoz Gonzalez, M., Penacchio, O., Raducanu, B., & Otazu, X. (2024). *Aesthetics without semantics* (SSRN Scholarly Paper No. 4817083). Social Science Research Network. https://doi.org/10.2139/ssrn.4817083

Pelowski, M., Markey, P. S., Forster, M., Gerger, G., & Leder, H. (2017). Move me, astonish me… delight my eyes and brain: The Vienna Integrated Model of top-down and bottom-up processes in Art Perception (VIMAP) and corresponding affective, evaluative, and neurophysiological correlates. *Physics of Life Reviews*, *21*, 80–125. https://doi.org/10.1016/j.plrev.2017.02.003

Pin, S. H. D., & Amirshahi, S. A. (2025). Individual differences in subjective image quality assessment: A Bayesian mixed-effects modeling approach. *IEEE Access*, *13*, 127504–127517. https://doi.org/10.1109/ACCESS.2025.3579016

Pombo, M., Igdalova, A., & Pelli, D. G. (2024). Consensus and contention in beauty judgment. *iScience*, *27*(7), 110213. https://doi.org/10.1016/j.isci.2024.110213

Reber, R., Schwarz, N., & Winkielman, P. (2004). Processing fluency and aesthetic pleasure: Is beauty in the perceiver's processing experience? *Personality and Social Psychology Review*, *8*(4), 364–382. https://doi.org/10.1207/S15327957PSPR0804_3

Redies, C. (2015). Combining universal beauty and cultural context in a unifying model of visual aesthetic experience. *Frontiers in Human Neuroscience*, *9*. https://doi.org/10.3389/fnhum.2015.00218

Redies, C., Bartho, R., Koßmann, L., Spehar, B., Hübner, R., Wagemans, J., & Hayn-Leichsenring, G. U. (2025). A toolbox for calculating quantitative image properties in aesthetics research. *Behavior Research Methods*, *57*(4), 117. https://doi.org/10.3758/s13428-025-02632-3

Ren, J., Shen, X., Lin, Z., Mech, R., & Foran, D. J. (2017). Personalized image aesthetics. *Proceedings of the IEEE International Conference on Computer Vision*, *2017-October*, 638–647. https://doi.org/10.1109/ICCV.2017.76

Rubio, F., Flores, M. J., & Puerta, J. M. (2022). Ranking-based scores for the assessment of aesthetic quality in photography. *Signal Processing: Image Communication*, *108*, 116803. https://doi.org/10.1016/J.IMAGE.2022.116803

Schloss, K. B., & Palmer, S. E. (2011). Aesthetic response to color combinations: Preference, harmony, and similarity. *Attention, Perception, & Psychophysics*, *73*(2), 551–571. https://doi.org/10.3758/s13414-010-0027-0

Schwarz, K., Wieschollek, P., & Lensch, H. P. A. (2018). Will people like your image? Learning the aesthetic space. *Proceedings - 2018 IEEE Winter Conference on Applications of Computer Vision, WACV 2018*, *2018-January*, 2048–2057. https://doi.org/10.1109/WACV.2018.00226

Segalin, C., Perina, A., Cristani, M., & Vinciarelli, A. (2017). The pictures we like are our image: Continuous mapping of favorite pictures into self-assessed and attributed personality traits. *IEEE Transactions on Affective Computing*, *8*(2), 268–285. IEEE Transactions on Affective Computing. https://doi.org/10.1109/TAFFC.2016.2516994

Skov, M., & Nadal, M. (2020). There are no aesthetic emotions: Comment on Menninghaus et al. (2019). *Psychological Review*, *127*(4), 640–649. https://doi.org/10.1037/rev0000187

Spehar, B., Walker, N., & Taylor, R. P. (2016). Taxonomy of individual variations in aesthetic responses to fractal patterns. *Frontiers in Human Neuroscience*, *10*, 350. https://doi.org/10.3389/fnhum.2016.00350

Srinivasa Desikan, B., Shimao, H., & Miton, H. (2022). WikiArtVectors: Style and color representations of artworks for cultural analysis via information theoretic measures. *Entropy*, *24*(9), 1175. https://doi.org/10.3390/e24091175

Sun, M., & Ying, H. (2023). Color’s perceptual diversity and categorical harmony improve aesthetic experience. *Psychology of Aesthetics, Creativity, and the Arts*. https://doi.org/10.1037/aca0000583

Sun, W.-T., Chao, T.-H., Kuo, Y.-H., & Hsu, W. H. (2017). Photo filter recommendation by category-aware aesthetic learning. *IEEE Transactions on Multimedia*, *19*(8), 1870–1880. IEEE Transactions on Multimedia. https://doi.org/10.1109/TMM.2017.2688929

Sun, Z., & Firestone, C. (2022). Beautiful on the inside: Aesthetic preferences and the skeletal complexity of shapes. *Perception*, *51*(12), 904–918. https://doi.org/10.1177/03010066221124872

The European Union Intellectual Property Office. (2025, January 13). *Copyright: Artworks entering the public domain in 2025*. EUIPO. https://www.euipo.europa.eu/en/news/artworks-entering-the-public-domain-in-2025

Van Geert, E., Bossens, C., & Wagemans, J. (2022). The Order & Complexity Toolbox for Aesthetics (OCTA): A systematic approach to study the relations between order, complexity, and aesthetic appreciation. *Behavior Research Methods*, 1–24. https://doi.org/10.3758/S13428-022-01900-W/FIGURES/17

Van Geert, E., & Wagemans, J. (2020). Order, complexity, and aesthetic appreciation. *Psychology of Aesthetics, Creativity, and the Arts*, *14*(2), 135–154. https://doi.org/10.1037/ACA0000224

Van Geert, E., Warny, A., & Wagemans, J. (2025). A systematic approach to study preferences for complexity at different levels of order. *Psychology of Aesthetics, Creativity, and the Arts*. https://doi.org/10.1037/aca0000816

van Zuijlen, M. J. P., Lin, H., Bala, K., Pont, S. C., & Wijntjes, M. W. A. (2020). *Materials In Paintings (MIP): An interdisciplinary dataset for perception, art history, and computer vision.* arXiv. https://doi.org/10.48550/arXiv.2012.02996

Vartanian, O., Navarrete, G., Chatterjee, A., Fich, L. B., Leder, H., Modrono, C., Nadal, M., Rostrup, N., & Skov, M. (2013). Impact of contour on aesthetic judgments and approach-avoidance decisions in architecture. *Proceedings of the National Academy of Sciences of the United States of America*, *110*(SUPPL2), 10446–10453. https://doi.org/10.1073/PNAS.1301227110/ASSET/09FE599C-C540-4DB1-A29A-5B3235B793BA/ASSETS/GRAPHIC/PNAS.1301227110FIG06.JPEG

Wang, J., Duan, H., Liu, J., Chen, S., Min, X., & Zhai, G. (2024). AIGCIQA2023: A large-scale image quality assessment database for AI generated images: From the perspectives of quality, authenticity and correspondence. In L. Fang, J. Pei, G. Zhai, & R. Wang (Eds.), *Artificial Intelligence* (pp. 46–57). Springer Nature. https://doi.org/10.1007/978-981-99-9119-8_5

Wang, Z., & Bovik, A. C. (2006). Introduction. In Z. Wang & A. C. Bovik (Eds.), *Modern Image Quality Assessment* (pp. 1–16). Springer International Publishing. https://doi.org/10.1007/978-3-031-02238-8_1

Wang, Z., Bovik, A. C., Sheikh, H. R., & Simoncelli, E. P. (2004). Image quality assessment: From error visibility to structural similarity. *IEEE Transactions on Image Processing*, *13*(4), 600–612. https://doi.org/10.1109/TIP.2003.819861

*WikiArt.org—Visual Art Encyclopedia*. (n.d.). Www.Wikiart.Org. Retrieved February 27, 2026, from https://www.wikiart.org/en/about

Wilson, A., & Chatterjee, A. (2005). The assessment of preference for balance: Introducing a new test. *Empirical Studies of the Arts*, *23*(2), 165–180. https://doi.org/10.2190/B1LR-MVF3-F36X-XR64

World Intellectual Property Organization. (n.d.). *Berne Convention for the Protection of Literary and Artistic Works*. Retrieved September 26, 2025, from https://www.wipo.int/treaties/en/ip/berne/index.html

Yi, R., Tian, H., Gu, Z., Lai, Y.-K., & Rosin, P. L. (2023). Towards artistic image aesthetics assessment: A large-scale dataset and a new method. *Proceedings of the IEEE/CVF Conference on Computer Vision and Pattern Recognition (CVPR),* 22388–22397. https://doi.org/10.1109/cvpr52729.2023.02144

Zhang, J., Miao, Y., & Yu, J. (2021). A comprehensive survey on computational aesthetic evaluation of visual art images: Metrics and challenges. *IEEE Access*, *9*, 77164–77187. https://doi.org/10.1109/ACCESS.2021.3083075

Zhong, Z., Zhou, F., & Qiu, G. (2022). *Aesthetically relevant image captioning* (arXiv:2211.15378). arXiv. https://doi.org/10.48550/arXiv.2211.15378

Zhou, X., Jin, X., Lv, J., Huang, H., Mao, M., & Cui, S. (2022). *Aesthetic attributes assessment of Images with AMANv2 and DPC-CaptionsV2* (arXiv:2208.04522). arXiv. https://doi.org/10.48550/arXiv.2208.04522

## Supplementary Materials

**Table 1**

*DODA: The Current State of Dataset Collection*

| Name | Author |
|---|---|
| AAA: Assessment of Art Attributes | (Chatterjee et al., 2010) |
| AADB : Aesthetics with Attributes Database | (Kong et al., 2016) |
| Accounting for taste: Individual differences in preference for harmony | (Palmer & Griscom, 2013) |
| AesFeedback | (Y. Huang, Sheng, et al., 2024) |
| AesMMIT | (Y. Huang, Sheng, et al., 2024) |
| Aesthetic Preferences Among Spatial Patterns | (L. Huang, 2025) |
| Aesthetic Responses to Exact Fractals Driven by Physical Complexity | (Bies et al., 2016) |
| AGIQA-1K | (Z. Zhang et al., 2023) |
| AGIQA-3K | (Li et al., 2023) |
| AIGCIQA2023 | (J. Wang et al., 2024) |
| AMD-A: Aesthetic Mixed Dataset With Attributes | (Jin et al., 2022) |
| APB Dataset | (Wilson & Chatterjee, 2005) |
| APDD: Aesthetics of Paintings and Drawings Dataset | (Jin, Qiao, Lu, Gao, et al., 2024) |
| APDDv2: Aesthetics of Paintings and Drawings Dataset with Artist Labeled Scores and Comments | (Jin, Qiao, Lu, Wang, et al., 2024) |
| AROD: Aesthetic Ratings from Online Data | (Schwarz et al., 2018) |
| art.pics | (Thieleking et al., 2020) |
| ART500K / DeepArt | (Mao et al., 2017) |
| ArtBrench-10 | (Liao et al., 2022) |
| ArtDL | (Milani & Fraternali, 2021) |
| ArtELingo | (Mohamed et al., 2022) |
| ArtELingo-28 | (Mohamed et al., 2024) |
| ArtEmis: Affective Language for Visual Art | (Achlioptas et al., 2021) |
| ArtEmis v2 | (Mohamed et al., 2022) |
| ArtiMuse-10k | (Cao et al., 2025) |
| AVA: A large-scale database for aesthetic visual analysis | (Murray et al., 2012) |
| AVA-Captions | (Ghosal et al., 2019) |
| AVA-PD (Photographer Demographic) | (Kairanbay et al., 2019) |
| AVA-review dataset | (W. Wang et al., 2019) |
| BAID: Boldbrush Artistic Image Dataset | (Yi et al., 2023) |
| BAM! Behance Artistic Media Dataset | (Wilber et al., 2017) |
| Beautiful on the inside: Aesthetic preferences and the skeletal complexity of shapes. | (Z. Sun & Firestone, 2022) |
| Beauty in abstract paintings | (Mallon et al., 2014) |
| CADB: Image Composition Assessment Dataset | (B. Zhang et al., 2021) |
| Classifying paintings by artistic genre: An analysis of features & classifiers | (Zujovic et al., 2009) |
| Combining multiple kernels for efficient image classification | (Siddiquie et al., 2009) |
| Complexity and Structure in Visual Patterns | (Chipman, 1977) |
| Computer Analysis of Art | (Shamir & Tarakhovsky, 2012) |
| contempArt | (Huckle et al, 2020) |
| Contour-Architecture Dataset | (Vartanian et al., 2013) |
| CrowdBeauty / Hidden Beauty | (Schifanella et al., 2015) |
| Curvature or angularity | (Bertamini et al., 2016) |
| CVD Preference Database | (Chen et al., 2025) |
| DEArt: Dataset of European Art | (Reshetnikov et al., 2025) |
| DELAUNAY | (Gontier et al., 2022) |
| deviantArt (devart) Database | (Sartori et al., 2015) |
| DPC-Captions | (Jin et al., 2019) |
| DPC-CaptionsV2 | (Zhou et al., 2022) |
| DPChallenge.com Dataset | (Datta et al., 2008) |
| DRAM: Diverse Realism in Art Movements dataset | (Cohen et al., 2022) |
| D-ViSA: A Dataset for Detecting Visual Sentiment from Art Images | (Kim et al., 2023) |
| EAPD: Expert-labeled Aesthetics Perception Database | (Y. Huang, Yuan, et al., 2024) |
| Eisenman & Grove Polygon Preference and Creativity Study | (Eisenman & Grove, 1972) |
| EVA: Explainable Visual Aesthetics quality assessment Dataset | (Kang et al., 2020) |
| Evaluating Abstract Art | (Lyssenko et al., 2016) |
| FACD: Filter Aesthetic Comparison Dataset | (W.-T. Sun et al., 2017) |
| Flickr cropping dataset | (Chen et al., 2017) |
| FLICKR- Style | (Karayev et al, 2014) |
| FLICKR_AES | (Ren et al., 2017) |
| GDP: Gourmet Photography Dataset | (Sheng et al., 2018) |
| IAD: image aesthetics dataset | (Lu et al., 2015) |
| ICAA17K | (He et al., 2023) |
| IEA: Images with Aesthetics and Emotions (also UEA: Unified Aesthetic and Emotional Dataset) | (Yu et al., 2019) |
| Impressionism, Expressionism, Surrealism: Automated Recognition of Painters and Schools of Art | (Shamir et al, 2010) |
| Impressions | (Kruk et al., 2023) |
| Influence of Six Types of Visual Structure on Complexity Judgments in Children and Adults | (Chipman & Mendelson, 1979) |
| IUAV image dataset | (Bertamini & Sinico, 2021) |
| JenAesthetics Subjective Dataset | (Amirshahi et al., 2015) |
| JenAesthetics β | (Amirshahi et al., 2016) |
| LAION-Aesthetics | (LAION Project, 2022) |
| LAPIS: Leuven Artworks Personalized Image Set | (Maerten et al., 2025) |
| Liking versus Complexity: Decomposing the Inverted U-curve | (Güçlütürk et al., 2016) |
| LOAD: Leuven Orthogonalized Art Data Set | (Lin et al., 2025) |
| MART: Museum of Modern and Contemporary Art of Trento and Rovereto | (Yanulevskaya et al., 2012) |
| Materials in Paintings (MIP) | (van Zuijlen et al., 2022) |
| MIRFLICKR | (Huiskes & Lew, 2008) |
| MPAAD: Portrait attribute dataset | (H. Huang et al., 2022) |
| MSC: Minimum Semantic Content (MSC) image dataset | (Parraga et al., 2024) |
| MultitaskPaintings100k | (Bianco et al, 2018) |
| NeatlyOrganized1 | (Van Geert & Wagemans, 2021) |
| Novel Graphic Patterns Dataset | (Jacobsen & Höfel, 2002) |
| OASIS-Beauty | (Brielmann & Pelli, 2019) |
| Painter by numbers | (Nichol & Kan, 2016) |
| Paintings-91 | (Khan et al., 2014) |
| Pandora | (Florea et al., 2016) |
| PARA: Personalized Image Aesthetics Assessment With Rich Attributes | (Yang et al., 2022) |
| PCCD: Photo Critique Captioning Dataset | (K. Y. Chang et al., 2017) |
| Photo.net | (Datta et al., 2006) |
| PICD: Photographic Image Composition Dataset | (Zhao et al., 2025) |
| Pick-a-Pic Dataset | (Kirstain et al., 2023) |
| Preference for Randomized Visual Patterns | (Friedenberg, 2019) |
| Princeton Adobe Photo Triage | (H. Chang et al., 2016) |
| PRINTART | (Carneiro et al., 2012) |
| PsychoFlickr | (Segalin et al., 2017) |
| REAL-CUR | (Ren et al., 2017) |
| Relating Objective Complexity, Subjective Complexity and Beauty in Binary Pixel Patterns | (Nath et al., 2024) |
| Rijksmuseum dataset | (Mensink & van Gemert, 2014) |
| RPCD: Reddit Photo Critique Dataset | (Vera Nieto et al., 2022) |
| Simulacra Aesthetic Caption | (Pressman, 2022/2026) |
| Stochastic modeling western paintings for effective classification | (Shen, 2009) |
| StyleBabel | (Ruta et al., 2022) |
| Stylebreeder | (Zheng et al., 2024) |
| Symm2020 | (Damiano et al., 2023) |
| Symmetry and Meaningfulness in the Spotlight of Expertness | (Palko-Arndt et al., 2025) |
| TAD66K: Rethinking Image Aesthetics Assessment: Models, Datasets and Benchmarks | (He et al., 2022) |
| Taxonomy of Individual Variations in Aesthetic Responses to Fractal Patterns | (Spehar et al., 2016) |
| The Penn Center for Neuroaesthetics Artwork Repository | (Estrada Gonzalez et al., 2025) |
| The Photographer Eye: Teaching Multimodal Large Language Models to Understand Image Aesthetics like Photographers | (Qi et al., 2025) |
| The small step toward asymmetry: Aesthetic judgment of broken symmetries | (Gartus & Leder, 2013) |
| Unsplash Lite Dataset 1.2.2 | (Unsplash, 2014) |
| Van Gogh and Kröller-Müller Museums Dataset | (Johnson, et al., 2008) |
| VAPS: Vienna Art Picture System | (Fekete et al., 2022) |
| ViCo: Visual Contour Stimulus Set | (Clemente et al., 2023) |
| Waterloo IAA | (Liu & Wang, 2017) |
| Web Gallery of Art | (Krén & Marx, 1996) |
| WikiArt | (Saleh & Elgammal, 2015) |
| WikiArt Emotions | (Mohammad & Kiritchenko, 2018) |
| WikiArt Refined | (Tan et al., 2019) |
| WikiArt Vectors | (Desikan et al., 2022) |
| Wikipaintings | (Karayev et al., 2014) |

*Note*: Datasets and their authors listed in DODA as of June 2026.

**Table 2**

*Museums with Open-Source Online Collections*

| Museum Name | Link to Collection |
|---|---|
| Art Institute of Chicago | https://www.artic.edu/collection |
| Auckland War Memorial Museum | https://www.aucklandmuseum.com/discover/collections-online |
| British Museum | https://www.britishmuseum.org/collection/ |
| Brooklyn Museum | https://www.brooklynmuseum.org/search/ |
| Cleveland Museum of Art | https://www.clevelandart.org/art/collection/search |
| ColBase (National Institutes for Cultural Heritage of Japan) | https://colbase.nich.go.jp/?locale=en |
| Cooper Hewitt | https://www.cooperhewitt.org/ |
| Harvard Art Museums | https://harvardartmuseums.org/collections |
| Kansallisgalleria (Finnish National Gallery) | https://kokoelma.kansallisgalleria.fi/en |
| Louvre | https://collections.louvre.fr/en/ |
| Mauritshuis | https://www.mauritshuis.nl/en/our-collection |
| Minneapolis Institute of Art | https://new.artsmia.org/art-artists/explore |
| Museo Nacional Del Prado | https://www.museodelprado.es/en/the-collection |
| National Gallery (London) | https://www.nationalgallery.org.uk |
| National Gallery of Art | https://www.nga.gov/artworks/free-images-and-open-access |
| National Palace Museum (Taiwan) | https://digitalarchive.npm.gov.tw/opendata/ |
| Paris Musées | https://parismuseescollections.paris.fr/en/recherche |
| QAGOMA (Queensland, Australia) | https://collection.qagoma.qld.gov.au/ |
| Rijksmuseum | https://www.rijksmuseum.nl/en/collection |
| Science Museum Group (UK) | https://collection.sciencemuseumgroup.org.uk/ |
| Smithsonian Open Access | https://www.si.edu/openaccess |
| SMK (National Gallery of Denmark) | https://open.smk.dk/art?q=* |
| Te Papa (New Zealand) | https://collections.tepapa.govt.nz/ |
| The Cleveland Museum of Art | https://www.clevelandart.org/art/collection/search |
| The Getty | https://www.getty.edu/art/collection/ |
| The MET | https://www.metmuseum.org/art/collection |
| The Walters Art Museum | https://art.thewalters.org/browse/category/ |
| Victoria and Albert Museum | https://www.vam.ac.uk/ |
| Whitney Museum | https://whitney.org/collection/works |
| Yale Center for British Art | https://collections.britishart.yale.edu/ |

*Note*: Non-exhaustive list of online museum collections. Part of this overview was taken from Maarten Wijntjes Website (*Maarten Wijntjes*, n.d.) (5 sources) and Nikolai Huckle's Git Hub Repository *art-datasets* (Huckle, 2020a) (22 sources).

**References for Supplementary Material**

Achlioptas, P., Ovsjanikov, M., Haydarov, K., Elhoseiny, M., & Guibas, L. (2021). ArtEmis: Affective language for visual art. *Proceedings of the IEEE Computer Society Conference on Computer Vision and Pattern Recognition*, 11564–11574. https://doi.org/10.1109/CVPR46437.2021.01140

Amirshahi, S. A., Hayn-Leichsenring, G. U., Denzler, J., & Redies, C. (2015). JenAesthetics Subjective Dataset: Analyzing paintings by subjective scores. *Lecture Notes in Computer Science*, *8925*, 3–19. https://doi.org/10.1007/978-3-319-16178-5_1

Amirshahi, S. A., Hayn-Leichsenring, G. U., Denzler, J., & Redies, C. (2016, September 19). *Color: A crucial factor for aesthetic quality assessment in a subjective dataset of paintings*. 12th Congress of the International Colour Association (AIC). https://doi.org/10.48550/arXiv.1609.05583

Bertamini, M., Palumbo, L., Gheorghes, T. N., & Galatsidas, M. (2016). Do observers like curvature or do they dislike angularity? *British Journal of Psychology*, *107*(1), 154–178. https://doi.org/10.1111/BJOP.12132

Bertamini, M., & Sinico, M. (2021). A Study of objects with smooth or sharp features created as line drawings by individuals trained in design. *Empirical Studies of the Arts*, *39*(1), 61–77. https://doi.org/10.1177/0276237419897048

Bianco, S., Mazzini, D., Napoletano, P., & Schettini, R. (2019). Multitask painting categorization by deep multibranch neural network. *Expert Systems with Applications, 135*, 90–101. https://doi.org/10.1016/j.eswa.2019.05.036

Bies, A. J., Blanc-Goldhammer, D. R., Boydston, C. R., Taylor, R. P., & Sereno, M. E. (2016). Aesthetic responses to exact fractals driven by physical complexity. *Frontiers in Human Neuroscience*, *10*. https://doi.org/10.3389/fnhum.2016.00210

Brielmann, A. A., & Pelli, D. G. (2019). Intense beauty requires intense pleasure. *Frontiers in Psychology*, *10*. https://doi.org/10.3389/fpsyg.2019.02420

Cao, S., Ma, N., Li, J., Li, X., Shao, L., Zhu, K., Zhou, Y., Pu, Y., Wu, J., Wang, J., Qu, B., Wang, W., Qiao, Y., Yao, D., & Liu, Y. (2025). *ArtiMuse: Fine-grained image aesthetics assessment with joint scoring and expert-level understanding* (arXiv:2507.14533). arXiv. https://doi.org/10.48550/arXiv.2507.14533

Carneiro, G., da Silva, N. P., Del Bue, A., & Costeira, J. P. (2012). Artistic image classification: An analysis on the PRINTART database. In A. Fitzgibbon, S. Lazebnik, P. Perona, Y. Sato, & C. Schmid (Eds.), *Computer Vision – ECCV 2012* (pp. 143–157). Springer. https://doi.org/10.1007/978-3-642-33765-9_11

Chang, H., Yu, F., Wang, J., Ashley, D., & Finkelstein, A. (2016). Automatic triage for a photo series. *ACM Trans. Graph.*, *35*(4), 148:1-148:10. https://doi.org/10.1145/2897824.2925908

Chang, K. Y., Lu, K. H., & Chen, C. S. (2017). Aesthetic critiques generation for photos. *Proceedings of the IEEE International Conference on Computer Vision*, *2017-October*, 3534–3543. https://doi.org/10.1109/ICCV.2017.380

Chatterjee, A., Widick, P., Sternschein, R., Smith, W., & Bromberger, B. (2010). The assessment of art attributes. *Empirical Studies of the Arts*, *28*(2), 207–222. https://doi.org/10.2190/EM.28.2.f

Chen, Y.-L., Huang, T.-W., Chang, K.-H., Tsai, Y.-C., Chen, H.-T., & Chen, B.-Y. (2017). Quantitative analysis of automatic image cropping algorithms: A dataset and comparative study. *2017 IEEE Winter Conference on Applications of Computer Vision (WACV)*, 226–234. https://doi.org/10.1109/WACV.2017.32

Chen, L., Zhu, Z., Chen, X., Go, K., & Mao, X. (2025). Personalized image preference assessment for individuals with color vision deficiency. *2025 Nicograph International (NICOInt)*, 52–58. https://doi.org/10.1109/NICOInt67466.2025.00018

Chipman, S. F. (1977). Complexity and structure in visual patterns. *Journal of Experimental Psychology: General*, *106*(3), 269–301. https://doi.org/10.1037/0096-3445.106.3.269

Chipman, S. F., & Mendelson, M. J. (1979). Influence of six types of visual structure on complexity judgments in children and adults. *Journal of Experimental Psychology: Human Perception and Performance*, *5*(2), 365–378. https://doi.org/10.1037/0096-1523.5.2.365

Clemente, A., Penacchio, O., Vila-Vidal, M., Pepperell, R., & Ruta, N. (2023). Explaining the Curvature Effect: Perceptual and hedonic evaluations of visual contour.

*Psychology of Aesthetics, Creativity, and the Arts*. https://doi.org/10.1037/ACA0000561

Cohen, N., Newman, Y., & Shamir, A. (2022). *Semantic Segmentation in Art Paintings* (arXiv:2203.03238). arXiv. https://doi.org/10.48550/arXiv.2203.03238

Damiano, C., Wilder, J., Zhou, E. Y., Walther, D. B., & Wagemans, J. (2023). The role of local and global symmetry in pleasure, interest, and complexity judgments of natural scenes. *Psychology of Aesthetics, Creativity, and the Arts*, *17*(3), 322–337. https://doi.org/10.1037/aca0000398

Datta, R., Joshi, D., Li, J., & Wang, J. Z. (2006). Studying aesthetics in photographic images using a computational approach. In A. Leonardis, H. Bischof, & A. Pinz (Eds.), *Computer Vision – ECCV 2006* (pp. 288–301). Springer. https://doi.org/10.1007/11744078_23

Datta, R., Li, J., & Wang, J. Z. (2008). Algorithmic inferencing of aesthetics and emotion in natural images: An exposition. *2008 15th IEEE International Conference on Image Processing*, 105–108. https://doi.org/10.1109/ICIP.2008.4711702

Desikan, B. S., Shimao, H., & Miton, H. (2022). WikiArtVectors: Style and color representations of artworks for cultural analysis via information theoretic measures. *Entropy, 24*(9), 1175. https://doi.org/10.3390/e24091175

Eisenman, R., & Grove, M. S. (1972). Self-ratings of creativity, semantic differential ratings, and preferences for polygons varying in complexity, simplicity, and symmetry. *The Journal of Psychology*, *81*(1), 63–67. https://doi.org/10.1080/00223980.1972.9923789

Estrada Gonzalez, V., Bobrow, I., Cardillo, E. R., Kim, O., Meletaki, V., & Chatterjee, A. (2025). A normed art database that incorporates diverse cultures and genres:

The Penn Center for Neuroaesthetics artwork repository. *Psychology of Aesthetics, Creativity, and the Arts*. https://doi.org/10.1037/aca0000755

Fang, Y., Zhu, H., Zeng, Y., Ma, K., & Wang, Z. (2020). Perceptual quality assessment of smartphone photography. *2020 IEEE/CVF Conference on Computer Vision and Pattern Recognition (CVPR)*, 3674–3683. https://doi.org/10.1109/CVPR42600.2020.00373

Fekete, A., Pelowski, M., Specker, E., Brieber, D., Rosenberg, R., & Leder, H. (2022). The Vienna Art Picture System (VAPS): A data set of 999 paintings and subjective ratings for art and aesthetics research. *Psychology of Aesthetics, Creativity, and the Arts*, *17*(5), 660–671. https://doi.org/10.1037/ACA0000460

Florea, C., Condorovici, R., Vertan, C., Boia, R., Florea, L., & Vranceanu, R. (2016). Pandora: Description of a painting database for art movement recognition with baselines and perspectives. In *2016 24th European Signal Processing Conference (EUSIPCO)* (pp. 918–922). IEEE. https://doi.org/10.1109/EUSIPCO.2016.7760382

Friedenberg, J. (2019). Beauty in the eye of the beholder: Individual differences in preference for randomized visual patterns. *Experimental Psychology*, *66*(2), 112–125. https://doi.org/10.1027/1618-3169/a000432

Gartus, A., & Leder, H. (2013). The small step toward asymmetry: Aesthetic judgment of broken symmetries. *I-Perception*, *4*(5), 361–364. https://doi.org/10.1068/i0588sas

Ghosal, K., Rana, A., & Smolic, A. (2019). Aesthetic image captioning from weakly-labelled photographs. *Proceedings - 2019 International Conference on*

*Computer Vision Workshop, ICCVW 2019*, 4550–4560. https://doi.org/10.1109/ICCVW.2019.00556

Gontier, C., Jordan, J., & Petrovici, M. A. (2022). *DELAUNAY: A dataset of abstract art for psychophysical and machine learning research*. arXiv. https://doi.org/10.48550/arXiv.2201.12123

Güçlütürk, Y., Jacobs, R. H. A. H., & Lier, R. van. (2016). Liking versus complexity: Decomposing the inverted U-curve. *Frontiers in Human Neuroscience*, *10*. https://doi.org/10.3389/fnhum.2016.00112

He, S., Ming, A., Li, Y., Sun, J., Zheng, S., & Ma, H. (2023). Thinking image color aesthetics assessment: Models, datasets and benchmarks. *2023 IEEE/CVF International Conference on Computer Vision (ICCV)*, 21781–21790. https://doi.org/10.1109/ICCV51070.2023.01996

He, S., Zhang, Y., Xie, R., Jiang, D., & Ming, A. (2022). Rethinking image aesthetics assessment: Models, datasets and benchmarks. *Proceedings of the Thirty-First International Joint Conference on Artificial Intelligence (IJCAI-22),* 942-948. https://doi.org/10.24963/ijcai.2022/132

Hosu, V., Lin, H., Sziranyi, T., & Saupe, D. (2020). KonIQ-10k: An ecologically valid database for deep learning of blind image quality assessment. *IEEE Transactions on Image Processing*, *29*, 4041–4056. https://doi.org/10.1109/TIP.2020.2967829

Huang, H., Jin, X., Li, X., Cui, S., & Xiao, C. (2022). Aesthetic evaluation of Asian and Caucasian photos with overall and attribute scores. *Computers and Electrical Engineering*, *103*, 108341. https://doi.org/10.1016/J.COMPELECENG.2022.108341

Huang, L. (2025). Aesthetic preferences among spatial patterns: Large-scale experiment, comprehensive exploration, and a three-component regularity-based model. *Psychology of Aesthetics, Creativity, and the Arts*. https://doi.org/10.1037/aca0000774

Huang, Y., Sheng, X., Yang, Z., Yuan, Q., Duan, Z., Chen, P., Li, L., Lin, W., & Shi, G. (2024). *AesExpert: Towards multi-modality foundation model for image aesthetics perception* (arXiv:2404.09624). arXiv. https://doi.org/10.48550/arXiv.2404.09624

Huang, Y., Yuan, Q., Sheng, X., Yang, Z., Wu, H., Chen, P., Yang, Y., Li, L., & Lin, W. (2024). *AesBench: An expert benchmark for multimodal large language models on image aesthetics perception* (arXiv:2401.08276). arXiv. https://doi.org/10.48550/arXiv.2401.08276

Huckle, N. [georgeblck]. (2020a). *Art-datasets* [Computer software]. GitHub. Retrieved June 26, 2026, from https://github.com/georgeblck/art-datasets

Huckle, N., Garcia, N. & Nakashima, Y. (2020b). Demographic influences on contemporary art with unsupervised style embeddings. In L. Leal-Taixé & S. Roth (Eds.), *Computer Vision – ECCV 2020 Workshops* (pp. 126–142). Springer. https://doi.org/10.1007/978-3-030-66096-3_10

Huiskes, M. J., & Lew, M. S. (2008). The MIR Flickr retrieval evaluation. *Proceedings of the 1st International ACM Conference on Multimedia Information Retrieval, MIR2008, Co-Located with the 2008 ACM International Conference on Multimedia, MM'08*, 39–43. https://doi.org/10.1145/1460096.1460104

Jacobsen, T., & Höfel, L. (2002). Aesthetic judgments of novel graphic patterns: Analyses of individual judgments. *Perceptual and Motor Skills*, *95*(3 PART 1), 755–766. https://doi.org/10.2466/PMS.2002.95.3.755

Jin, X., Li, X., Lou, H., Fan, C., Deng, Q., Xiao, C., Cui, S., & Singh, A. K. (2022). Aesthetic attribute assessment of images numerically on mixed multi-attribute datasets. *ACM Transactions on Multimedia Computing, Communications and Applications*, *18*(3). https://doi.org/10.1145/3547144

Jin, X., Qiao, Q., Lu, Y., Gao, S., Huang, H., & Li, G. (2024). *Paintings and drawings aesthetics assessment with rich attributes for various artistic categories* (arXiv:2405.02982). arXiv. https://doi.org/10.48550/arXiv.2405.02982

Jin, X., Qiao, Q., Lu, Y., Wang, H., Huang, H., Gao, S., Liu, J., & Li, R. (2024). APDDv2: Aesthetics of paintings and drawings dataset with artist labeled scores and comments. In A. Globerson, L. Mackey, D. Belgrave, A. Fan, U. Paquet, J. Tomczak, & C. Zhang (Eds.), *Advances in Neural Information Processing Systems* (Vol. 37, pp. 103064–103075). Curran Associates, Inc. https://doi.org/10.52202/079017-3274

Jin, X., Wu, L., Zhao, G., Li, X., Zhang, X., Ge, S., Zou, D., Zhou, B., & Zhou, X. (2019). Aesthetic attributes assessment of images. *Proceedings of the 27th ACM International Conference on Multimedia*, 311–319. https://doi.org/10.1145/3343031.3350970

Johnson, C. R., Hendriks, Ella, Berezhnoy, I. J., Brevdo, E., Hughes, S. M., Daubechies, I., Li, J., Postma, E., & Wang, J. Z. (2008). Image processing for artist identification. *IEEE Signal Processing Magazine, 25*(4), 37–48. https://doi.org/10.1109/MSP.2008.923513

Kairanbay, M., See, J., & Wong, L. K. (2019). Beauty is in the eye of the beholder. *ACM Transactions on Multimedia Computing, Communications, and Applications (TOMM)*, *15*(2s). https://doi.org/10.1145/3328993

Kang, C., Valenzise, G., & Dufaux, F. (2020). EVA: An explainable visual aesthetics dataset. *ATQAM/MAST 2020 - Proceedings of the Joint Workshop on Aesthetic and Technical Quality Assessment of Multimedia and Media Analytics for Societal Trends*. https://doi.org/10.1145/3423268.3423590

Karayev, S., Hertzmann, A., Trentacoste, M., Han, H., Winnemoeller, H., Agarwala, A., & Darrell, T. (2014). Recognizing image style. In *Proceedings of the British Machine Vision Conference 2014* (pp. 122.1–122.11). British Machine Vision Association. https://doi.org/10.5244/c.28.122

Khan, F. S., Beigpour, S., van de Weijer, J., & Felsberg, M. (2014). Painting-91: A large scale database for computational painting categorization. Machine Vision and Applications, 25(6), 1385–1397. https://doi.org/10.1007/s00138-014-0621-6

Kim, S., An, C., Cha, J., Kim, D., & Park, E. (2023). D-ViSA: A dataset for detecting visual sentiment from art images. *2023 IEEE/CVF International Conference on Computer Vision Workshops (ICCVW)*, 3043–3051. https://doi.org/10.1109/ICCVW60793.2023.00328

Kirstain, Y., Polyak, A., Singer, U., Matiana, S., Penna, J., & Levy, O. (2023). *Pick-a-Pic: An open dataset of user preferences for text-to-image generation* (arXiv:2305.01569). arXiv. https://doi.org/10.48550/arXiv.2305.01569

Kong, S., Shen, X., Lin, Z., Mech, R., & Fowlkes, C. (2016). Photo aesthetics ranking network with attributes and content adaptation. In B. Leibe, J. Matas, N. Sebe, &

M. Welling (Eds.), *14th European Conference on Computer Vision (ECCV 2016): 9905 LNCS* (pp. 662–679). Springer. https://doi.org/10.1007/978-3-319-46448-0_40

Krén, E., & Marx, D. (1996). Web Gallery of Art. https://www.wga.hu/index_welcome.html

Kruk, J., Ziems, C., & Yang, D. (2023). Impressions: Understanding visual semiotics and aesthetic impact. *EMNLP 2023 - 2023 Conference on Empirical Methods in Natural Language Processing, Proceedings*, 12273–12291.

LAION Project. (2022). *LAION-Aesthetics* (Version 1) [Dataset]. GitHub. https://github.com/LAION-AI/laion-datasets/blob/main/laion-aesthetic.md

Li, C., Zhang, Z., Wu, H., Sun, W., Min, X., Liu, X., Zhai, G., & Lin, W. (2023). *AGIQA-3K: An open database for AI-generated image quality assessment* (arXiv:2306.04717). arXiv. https://doi.org/10.48550/arXiv.2306.04717

Liao, P., Li, X., Liu, X., & Keutzer, K. (2022). *The ArtBench Dataset: Benchmarking generative models with artworks*. https://arxiv.org/abs/2206.11404v1

Lin, Y., Op de Beeck, H., & Wagemans, J. (2025). Leuven Orthogonalized Art Data Set (LOAD): A multidimensional art image set for aesthetic appreciation research. *Psychology of Aesthetics, Creativity, and the Arts*. https://doi.org/10.1037/aca0000791

Liu, W., & Wang, Z. (2017). A database for perceptual evaluation of image aesthetics. *2017 IEEE International Conference on Image Processing (ICIP)*, 1317–1321. https://doi.org/10.1109/ICIP.2017.8296495

Lu, X., Lin, Z., Jin, H., Yang, J., & Wang, J. Z. (2015). Rating image aesthetics using deep learning. *IEEE Transactions on Multimedia*, *17*(11), 2021–2034. https://doi.org/10.1109/TMM.2015.2477040

Lyssenko, N., Redies, C., & Hayn-Leichsenring, G. U. (2016). Evaluating abstract art: Relation between term usage, subjective ratings, image properties and personality traits. *Frontiers in Psychology*, *7*. https://doi.org/10.3389/fpsyg.2016.00973

*Maarten Wijntjes*. (n.d.). Maarten Wijntjes. Retrieved April 9, 2026, from https://maartenwijntjes.github.io/

Maerten, A.-S., Chen, L.-W., Winter, S. D., Bossens, C., & Wagemans, J. (2025). LAPIS: A novel dataset for personalized image aesthetic assessment. *Proceedings of the IEEE/CVF Conference on Computer Vision and Pattern Recognition (CVPR) Workshops*, 6302–6311. https://doi.org/10.48550/arXiv.2504.07670

Mallon, B., Redies, C., & Hayn-Leichsenring, G. U. (2014). Beauty in abstract paintings: Perceptual contrast and statistical properties. *Frontiers in Human Neuroscience*, *8*. https://doi.org/10.3389/fnhum.2014.00161

Mao, H., Cheung, M., & She, J. (2017). DeepArt: Learning joint representations of visual arts. In *Proceedings of the 25th ACM International Conference on Multimedia* (pp. 1183–1191). Association for Computing Machinery. https://doi.org/10.1145/3123266.3123405

Mensink, T., & van Gemert, J. (2014). The Rijksmuseum challenge: Museum-centered visual recognition. In *Proceedings of the 2014 International Conference on Multimedia Retrieval* (pp. 451–454). Association for Computing Machinery. https://doi.org/10.1145/2578726.2578791

Milani, F., & Fraternali, P. (2021). A dataset and a convolutional model for iconography classification in paintings. *Journal on Computing and Cultural Heritage, 14*(4), Article 46. https://doi.org/10.1145/3458885

Mohamed, Y., Abdelfattah, M., Alhuwaider, S., Li, F., Zhang, X., Church, K., & Elhoseiny, M. (2022a). ArtELingo: A million emotion annotations of WikiArt with emphasis on diversity over language and culture. In Y. Goldberg, Z. Kozareva, & Y. Zhang (Eds.), *Proceedings of the 2022 Conference on Empirical Methods in Natural Language Processing* (pp. 8770–8785). Association for Computational Linguistics. https://doi.org/10.18653/v1/2022.emnlp-main.600

Mohamed, Y., Khan, F. F., Haydarov, K., & Elhoseiny, M. (2022b). It is okay to not be okay: Overcoming emotional bias in affective image captioning by contrastive data collection. *Proceedings of the IEEE Computer Society Conference on Computer Vision and Pattern Recognition*, *2022-June*, 21231–21240. https://doi.org/10.1109/CVPR52688.2022.02058

Mohamed, Y., Li, R., Ahmad, I. S., Haydarov, K., Torr, P., Church, K., & Elhoseiny, M. (2024, November). No culture left behind: ArtELingo-28, a benchmark of WikiArt with captions in 28 languages. In Y. Al-Onaizan, M. Bansal, & Y.-N. Chen (Eds.), *Proceedings of the 2024 Conference on Empirical Methods in Natural Language Processing* (pp. 20939–20962). Association for Computational Linguistics. https://doi.org/10.18653/v1/2024.emnlp-main.1165

Mohammad, S., & Kiritchenko, S. (2018). WikiArt Emotions: An annotated dataset of emotions evoked by art. In N. Calzolari, K. Choukri, C. Cieri, T. Declerck, S. Goggi, K. Hasida, H. Isahara, B. Maegaard, J. Mariani, H. Mazo, A. Moreno, J. Odijk, S. Piperidis, & T. Tokunaga (Eds.), *Proceedings of the Eleventh*

*International Conference on Language Resources and Evaluation (LREC 2018)*. European Language Resources Association (ELRA). https://aclanthology.org/L18-1197/

Murray, N., Marchesotti, L., & Perronnin, F. (2012). AVA: A large-scale database for aesthetic visual analysis. *Proceedings of the IEEE Computer Society Conference on Computer Vision and Pattern Recognition*, 2408–2415. https://doi.org/10.1109/CVPR.2012.6247954

Nath, S. S., Brändle, F., Schulz, E., Dayan, P., & Brielmann, A. (2024). Relating objective complexity, subjective complexity, and beauty in binary pixel patterns. *Psychology of Aesthetics, Creativity, and the Arts*. https://doi.org/10.1037/aca0000657

Nichol, K. [small yellow duck], & Kan, W. (2016). Painter by numbers [Dataset]. Kaggle. https://kaggle.com/competitions/painter-by-numbers

Palko-Arndt, B., Bali, C., & Illes, A. (2025). Symmetry and meaningfulness in the spotlight of expertness. *Empirical Studies of the Arts*, *43*(2), 991–1012. https://doi.org/10.1177/02762374241291012

Palmer, S. E., & Griscom, W. S. (2013). Accounting for taste: Individual differences in preference for harmony. *Psychonomic Bulletin & Review*, *20*(3), 453–461. https://doi.org/10.3758/s13423-012-0355-2

Parraga, C. A., Muñoz Gonzalez, M., Penacchio, O., Raducanu, B., & Otazu, X. (2024). *Aesthetics without semantics* (SSRN Scholarly Paper No. 4817083). Social Science Research Network. https://doi.org/10.2139/ssrn.4817083

Pressman, J. D. (2026). *JD-P/simulacra-aesthetic-captions* [Computer software]. https://github.com/JD-P/simulacra-aesthetic-captions (Original work published 2022)

Qi, D., Zhao, H., Shi, J., Jenni, S., Fan, Y., Dernoncourt, F., Cohen, S., & Li, S. (2025). *The photographer eye: Teaching multimodal large language models to understand image aesthetics like photographers* (arXiv:2509.18582). arXiv. https://doi.org/10.48550/arXiv.2509.18582

Ren, J., Shen, X., Lin, Z., Mech, R., & Foran, D. J. (2017). Personalized image aesthetics. *Proceedings of the IEEE International Conference on Computer Vision*, *2017-October*, 638–647. https://doi.org/10.1109/ICCV.2017.76

Reshetnikov, A., Marinescu, M.-C., Lopez, J. M., Mendoza, S., Freire, N., Marrero, M., Tsoupra, E., & Isaac, A. (2025). DEArt: Building and evaluating a dataset for object detection and pose classification for European art. *Journal of Cultural Heritage, 75*, 258–266. https://doi.org/10.1016/j.culher.2025.07.022

Ruta, D., Gilbert, A., Aggarwal, P., Marri, N., Kale, A., Briggs, J., Speed, C., Jin, H., Faieta, B., Filipkowski, A., Lin, Z., & Collomosse, J. (2022). StyleBabel: Artistic style tagging and captioning. In S. Avidan, G. Brostow, M. Cissé, G. M. Farinella, & T. Hassner (Eds.), *Computer Vision – ECCV 2022* (pp. 219–236). Springer. https://doi.org/10.1007/978-3-031-20074-8_13

Saleh, B., & Elgammal, A. (2015). *Large-scale classification of fine-art paintings: Learning the right metric on the right feature* (arXiv:1505.00855). arXiv. https://doi.org/10.48550/arXiv.1505.00855

Sartori, A., Uijlings, J., Yanulevskaya, V., Salah, A. A., Bruni, E., Sebe, N., Salah, A. A., Hung, H., Aran, O., Gunes, H., & Turk, M. (2015). Affective analysis of professional and amateur abstract paintings using statistical analysis and art theory. *ACM Trans. Interact. Intell. Syst*, *5*(8). https://doi.org/10.1145/2768209

Schifanella, R., Redi, M., & Aiello, L. M. (2015). An image is worth more than a thousand favorites: Surfacing the hidden beauty of Flickr pictures. *Proceedings of the 9th International Conference on Web and Social Media, ICWSM 2015*, 397–406. https://doi.org/10.1609/icwsm.v9i1.14612

Schwarz, K., Wieschollek, P., & Lensch, H. P. A. (2018). Will people like your image? Learning the aesthetic space. *Proceedings - 2018 IEEE Winter Conference on Applications of Computer Vision, WACV 2018*, *2018-January*, 2048–2057. https://doi.org/10.1109/WACV.2018.00226

Segalin, C., Perina, A., Cristani, M., & Vinciarelli, A. (2017). The pictures we like are our image: Continuous mapping of favorite pictures into self-assessed and attributed personality traits. *IEEE Transactions on Affective Computing*, *8*(2), 268–285. IEEE Transactions on Affective Computing. https://doi.org/10.1109/TAFFC.2016.2516994

Shamir, L., Macura, T., Orlov, N., Eckley, D. M., & Goldberg, I. G. (2010). Impressionism, expressionism, surrealism: Automated recognition of painters and schools of art. *ACM Transactions on Applied Perception, 7*(2), Article 8. https://doi.org/10.1145/1670671.1670672

Shamir, L., & Tarakhovsky, J. A. (2012). Computer analysis of art. *Journal on Computing and Cultural Heritage, 5*(2), Article 7. https://doi.org/10.1145/2307723.2307726

Sheng, K., Dong, W., Huang, H., Ma, C., & Hu, B. G. (2018). Gourmet photography dataset for aesthetic assessment of food images. *SIGGRAPH Asia 2018 Technical Briefs, SA 2018*. https://doi.org/10.1145/3283254.3283260

Siddiquie, B., Vitaladevuni, S. N., & Davis, L. S. (2009). Combining multiple kernels for efficient image classification. *In 2009 Workshop on Applications of Computer Vision (WACV)* (pp. 1–8). IEEE. https://doi.org/10.1109/WACV.2009.5403040

Spehar, B., Walker, N., & Taylor, R. P. (2016). Taxonomy of individual variations in aesthetic responses to fractal patterns. *Frontiers in Human Neuroscience*, *10*. https://doi.org/10.3389/fnhum.2016.00350

Srinivasa Desikan, B., Shimao, H., & Miton, H. (2022). WikiArtVectors: Style and color representations of artworks for cultural analysis via information theoretic measures. *Entropy*, *24*(9), 1175. https://doi.org/10.3390/e24091175

Sun, W.-T., Chao, T.-H., Kuo, Y.-H., & Hsu, W. H. (2017). Photo filter recommendation by category-aware aesthetic learning. *IEEE Transactions on Multimedia*, *19*(8), 1870–1880. IEEE Transactions on Multimedia. https://doi.org/10.1109/TMM.2017.2688929

Sun, Z., & Firestone, C. (2022). Beautiful on the inside: Aesthetic preferences and the skeletal complexity of shapes. *Perception*, *51*(12), 904–918. https://doi.org/10.1177/03010066221124872

Tan, W. R., Chan, C. S., Aguirre, H. E., & Tanaka, K. (2019). Improved ArtGAN for conditional synthesis of natural image and artwork. *IEEE Transactions on Image Processing*, *28*(1), 394–409. https://doi.org/10.1109/TIP.2018.2866698

Tang, X., Luo, W., & Wang, X. (2013). Content-based photo quality assessment. *IEEE Transactions on Multimedia*, *15*(8), 1930–1943. https://doi.org/10.1109/TMM.2013.2269899

Thieleking, R., Medawar, E., Disch, L., & Witte, A. V. (2020). art.pics Database: An open access database for art stimuli for experimental research. *Frontiers in Psychology*, *11*. https://doi.org/10.3389/fpsyg.2020.576580

Unsplash. (2014). *Join Unsplash: Beautiful, free photos and illustrations*. https://unsplash.com/about

Van Geert, E., & Wagemans, J. (2021). Order, complexity, and aesthetic preferences for neatly organized compositions. *Psychology of Aesthetics, Creativity, and the Arts*, *15*(3), 484–504. https://doi.org/10.1037/aca0000276

van Zuijlen, M. J. P., Lin, H., Bala, K., Pont, S. C., & Wijntjes, M. W. A. (2020). *Materials In Paintings (MIP): An interdisciplinary dataset for perception, art history, and computer vision.* arXiv. https://doi.org/10.48550/arXiv.2012.02996

Vartanian, O., Navarrete, G., Chatterjee, A., Fich, L. B., Leder, H., Modrono, C., Nadal, M., Rostrup, N., & Skov, M. (2013). Impact of contour on aesthetic judgments and approach-avoidance decisions in architecture. *Proceedings of the National Academy of Sciences of the United States of America*, *110*(SUPPL2), 10446–10453. https://doi.org/10.1073/PNAS.1301227110/ASSET/09FE599C-C540-4DB1-A29A-5B3235B793BA/ASSETS/GRAPHIC/PNAS.1301227110FIG06.JPEG

Vera Nieto, D. V., Celona, L., & Fernandez-Labrador, C. (2022). *Understanding aesthetics with language: A photo critique dataset for aesthetic assessment* (arXiv:2206.08614). arXiv. https://doi.org/10.48550/arXiv.2206.08614

Wang, J., Duan, H., Liu, J., Chen, S., Min, X., & Zhai, G. (2024). AIGCIQA2023: A large-scale image quality assessment database for AI generated images: From the perspectives of quality, authenticity and correspondence. In L. Fang, J. Pei,

G. Zhai, & R. Wang (Eds.), *Artificial Intelligence* (pp. 46–57). Springer Nature. https://doi.org/10.1007/978-981-99-9119-8_5

Wang, W., Yang, S., Zhang, W., & Zhang, J. (2019). Neural aesthetic image reviewer. *IET Computer Vision*, *13*(8), 749–758. https://doi.org/10.1049/IET-CVI.2019.0361

Wilber, M. J., Fang, C., Jin, H., Hertzmann, A., Collomosse, J., & Belongie, S. (2017). BAM! The Behance Artistic Media Dataset for recognition beyond photography. In *Proceedings of the IEEE International Conference on Computer Vision (ICCV)* (pp. 1211–1220). IEEE. https://doi.org/10.1109/ICCV.2017.136

Wilson, A., & Chatterjee, A. (2005). The assessment of preference for balance: Introducing a new test. *Empirical Studies of the Arts*, *23*(2), 165–180. https://doi.org/10.2190/B1LR-MVF3-F36X-XR64

Yang, Y., Xu, L., Li, L., Qie, N., Li, Y., Zhang, P., & Guo, Y. (2022). *Personalized image aesthetics assessment with rich attributes* (arXiv:2203.16754). arXiv. https://doi.org/10.48550/arXiv.2203.16754

Yanulevskaya, V., Uijlings, J., Bruni, E., Sartori, A., Zamboni, E., Bacci, F., Melcher, D., & Sebe, N. (2012). In the eye of the beholder: Employing statistical analysis and eye tracking for analyzing abstract paintings. *Proceedings of the 20th ACM International Conference on Multimedia, MM ’12*, 349–358. https://doi.org/10.1145/2393347.2393399

Yi, R., Tian, H., Gu, Z., Lai, Y.-K., & Rosin, P. L. (2023). Towards artistic image aesthetics assessment: A large-scale dataset and a new method. *Proceedings of the IEEE/CVF Conference on Computer Vision and Pattern Recognition (CVPR),* 22388–22397. https://doi.org/10.1109/cvpr52729.2023.02144

Yu, J., Cui, C., Geng, L., Ma, Y., & Yin, Y. (2019). Towards unified aesthetics and emotion prediction in images. *2019 IEEE International Conference on Image Processing (ICIP)*, 2526–2530. https://doi.org/10.1109/ICIP.2019.8803388

Zhang, B., Niu, L., & Zhang, L. (2021). *Image composition assessment with saliency-augmented multi-pattern pooling* (arXiv:2104.03133). arXiv. https://doi.org/10.48550/arXiv.2104.03133

Zhang, Z., Li, C., Sun, W., Liu, X., Min, X., & Zhai, G. (2023). *A perceptual quality assessment exploration for AIGC images* (arXiv:2303.12618). arXiv. https://doi.org/10.48550/arXiv.2303.12618

Zhao, Z., Lu, P., Zhang, A., Li, P., Li, X., Liu, X., Hu, Y., Chen, S., Wang, L., & Guo, W. (2025). Can machines understand composition? Dataset and benchmark for photographic image composition embedding and understanding. *Proceedings of the IEEE/CVF Conference on Computer Vision and Pattern Recognition (CVPR)*, 14411–14421.

Zheng, M., Simsar, E., Yesiltepe, H., Tombari, F., Simon, J., & Yanardag, P. (2024). Stylebreeder: Exploring and democratizing artistic styles through text-to-image models. NeurIPS Proceedings. https://openreview.net/forum?id=EvgyfFsv0w

Zhong, Z., Zhou, F., & Qiu, G. (2022). *Aesthetically Relevant Image Captioning* (arXiv:2211.15378). arXiv. https://doi.org/10.48550/arXiv.2211.15378

Zhou, X., Jin, X., Lv, J., Huang, H., Mao, M., & Cui, S. (2022). *Aesthetic attributes assessment of images with AMANv2 and DPC-CaptionsV2* (arXiv:2208.04522). arXiv. https://doi.org/10.48550/arXiv.2208.04522

Zujovic, J., Gandy, L., Friedman, S., Pardo, B., & Pappas, T. N. (2009). Classifying paintings by artistic genre: An analysis of features & classifiers. In *2009 IEEE*

*International Workshop on Multimedia Signal Processing* (pp. 1–5). IEEE.
https://doi.org/10.1109/MMSP.2009.5293271

## CRediT Statement

**Lisa Koßmann:** Conceptualization, Data curation, Formal analysis, Investigation, Methodology, Project administration, Validation, Visualization, Writing - original draft, and Writing - review & editing.
**Ralf Bartho:** Conceptualization, Data curation, Formal analysis, Investigation, Methodology, Software, Validation, and Writing - review & editing.
**Christoph Redies:** Conceptualization, Resources, Supervision, and Writing - review & editing.
**Johan Wagemans:** Conceptualization, Funding acquisition, Resources, Supervision, and Writing - review & editing.